\documentclass{article}

\usepackage{microtype}
\usepackage{graphicx}
\usepackage{subfigure}
\usepackage{booktabs}

\usepackage{amsmath}
\usepackage{multirow}
\usepackage[titletoc]{appendix}
\usepackage{caption}

\usepackage{hyperref}

\usepackage[accepted]{icml2021}

\begin{document}

\twocolumn[
\icmltitle{SinIR: Efficient General Image Manipulation with Single Image Reconstruction}

\icmlsetsymbol{equal}{*}

\begin{icmlauthorlist}
\icmlauthor{Jihyeong Yoo}{hkust}
\icmlauthor{Qifeng Chen}{hkust}
\end{icmlauthorlist}

\icmlaffiliation{hkust}{Department of Computer Science and Engineering, The Hong Kong University of Science and Technology, Clear Water Bay, Hong Kong}

\icmlcorrespondingauthor{Qifeng Chen}{cqf@ust.hk}

\icmlkeywords{Machine Learning, ICML}

\vskip 0.3in
]

\printAffiliationsAndNotice{}

\begin{abstract}
We propose SinIR, an efficient reconstruction-based framework trained on a single natural image for general image manipulation, including super-resolution, editing, harmonization, paint-to-image, photo-realistic style transfer, and artistic style transfer. We train our model on a single image with cascaded multi-scale learning, where each network at each scale is responsible for image reconstruction. This reconstruction objective greatly reduces the complexity and running time of training, compared to the GAN objective. However, the reconstruction objective also exacerbates the output quality. Therefore, to solve this problem, we further utilize simple random pixel shuffling, which also gives control over manipulation, inspired by the Denoising Autoencoder. With quantitative evaluation, we show that SinIR has competitive performance on various image manipulation tasks. Moreover, with a much simpler training objective (\emph{i.e.}, reconstruction), SinIR is trained 33.5 times faster than SinGAN (for \(500 \times 500\) images) that solves similar tasks. Our code is publicly available at \href{https://github.com/YooJiHyeong/SinIR}{github.com/YooJiHyeong/SinIR}.
\end{abstract}


\begin{figure*}
\begin{center}
\includegraphics[width=\linewidth]{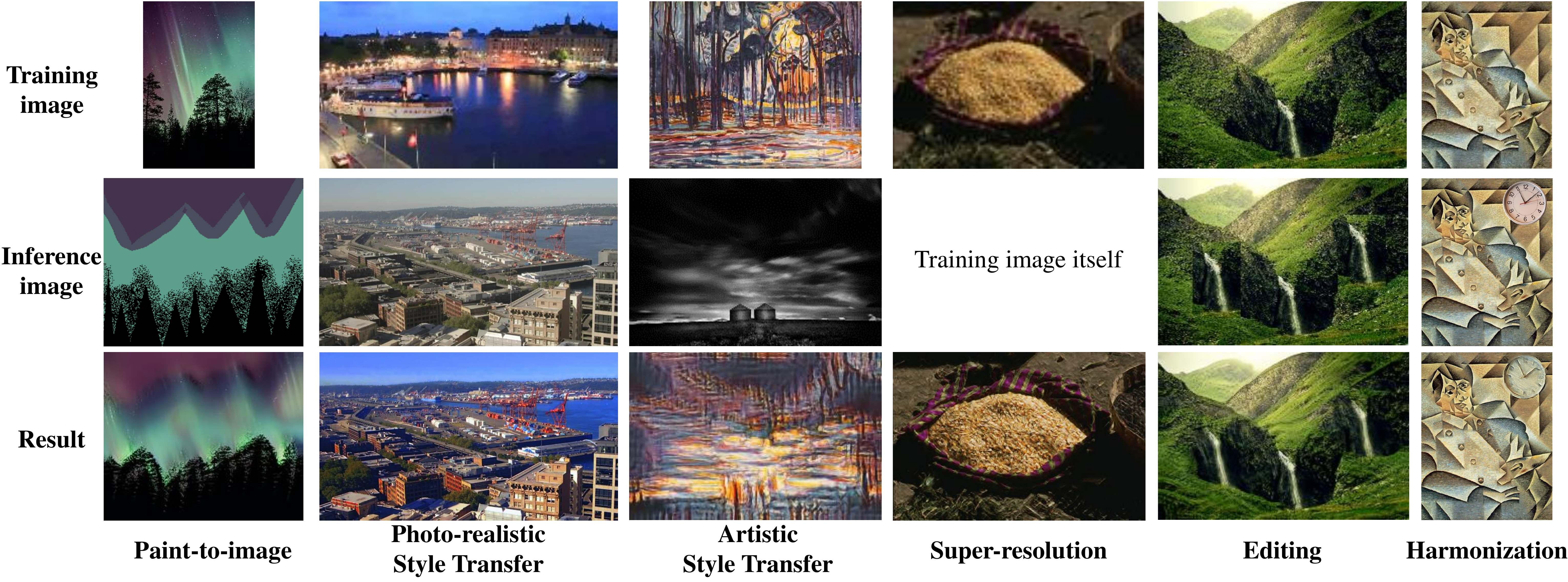}
\vspace{-0.6cm}
  \caption{\textbf{General image manipulation of SinIR.} SinIR is trained on a \emph{single natural image} with \emph{reconstruction loss}, cascaded multi-scale learning, and random pixel shuffling. Once trained, SinIR can manipulate any image utilizing internal information learned from the training image. Best viewed when zoomed.}
  \label{main}
\end{center}
\end{figure*}

\section{\label{sec1}Introduction}

Researchers in image processing have increasing interests in \emph{deep internal learning}, which can solve image manipulation problems by training a model on one single image and not relying on a large-scale dataset. Training on a single image is plausible as the statistics of a single image have abundant information that can be used as a powerful prior for solving various problems \cite{Shaham19}. Internal learning has long been employed in prior work even before deep learning became popular, proving internal approaches can be successfully applied to several image manipulation tasks (\emph{e.g.}, super-resolution \cite{Glasner09}, editing \cite{He12}, dehazing \cite{Bahat16}, texture synthesis \cite{Efros99}, and segmentation \cite{Bagon08}).

Recently, several \emph{deep internal learning} methods \cite{Shocher18zeroshot} are proposed and achieve remarkable performance that is comparable to that of external methods trained on large-scale datasets. For example, they solve super-resolution \cite{Shocher18zeroshot, Ulyanov18, Bell-Kligler19}, restoration \cite{Zhang19restor, MastanR20}, reflection removal \cite{Fan19}, deblur \cite{Ren19}, segmentation and dehazing \cite{Gandelsman19}, denoising, and inpainting \cite{Ulyanov18, Zhang19vidinpaint} tasks. Utilizing Generative Adversarial Networks (GAN) \cite{Goodfellow14}, several approaches \cite{Li16, Jetchev16, Bergmann17, Zhou018} solve texture generation from a single texture image. Expanding the capacity of GANs, InGAN \cite{Shocher19} and DCIL \cite{MastanR20} solve retargeting with a single natural image.

However, these methods have critical problems that prevent the practical application of deep internal learning. First, most of these methods are image-specific in terms of manipulation, except for MGANs \cite{Li16}. This means that the trained models can only manipulate the training images, and for other images, separate models have to be trained. Second, most of these methods are task-specific. This means that the trained models can only perform one specific image manipulation. Although DIP \cite{Ulyanov18}, Double-DIP \cite{Gandelsman19}, and DCIL \cite{MastanR20} can solve several tasks, these models can only conduct one specific manipulation at a time.

SinGAN \cite{Shaham19}, one of the most recent works in deep internal learning, was the first to achieve general image manipulation, where one trained model can solve various problems, including super-resolution, editing, paint-to-image, and harmonization. SinGAN learns unconditional generation (\emph{i.e.}, mapping noise to images, not images to images) and then uses the generative power for image manipulation. However, this leads to long training time (1.5 and 4.5 hours for \(250 \times 250\) and \(500 \times 500\) images on RTX 2080 Ti respectively. See Table \ref{analysis}). As unconditional image generation is a relatively difficult problem, SinGAN uses a sophisticated loss function (\emph{i.e.}, WGAN-GP loss \cite{Gulrajani17}) for better convergence and trains multiple GANs for a large number of iterations. Although SinGAN is not an image-specific and task-specific framework, this prolonged training time still hinders practical usage of deep internal learning.

In this work, we circumvent this problem with \textbf{SinIR}, a reconstruction-based framework trained on a single natural image for general image manipulation. Our model learns \textbf{image reconstruction}, which is a much simpler problem compared to the unconditional generation of SinGAN \cite{Shaham19}. This allows SinIR to achieve a vastly shorter training time (33.49 times faster than SinGAN for \(500 \times 500\) images). To this end, \textbf{cascaded multi-scale learning} is employed to learn robust cross-scale representations. However, due to the innate property of reconstruction, simply applying cascaded multi-scale learning leads to poor manipulation quality. Thus, inspired by denoising autoencoder \cite{Vincent08}, we introduce \textbf{random pixel shuffling} that effectively mitigates the problem without a significant increase in computational cost. Furthermore, we show that random pixel shuffling gives additional control over manipulation.

However, we want to make it clear that although SinIR provides a simpler solution to most of the tasks SinGAN can handle, SinIR cannot completely replace SinGAN. This is because SinIR and SinGAN show clearly different capabilities for some tasks. Particularly, SinIR shows less capability for random image generation but performs well on photo-realistic and artistic style transfer, and vice versa. To our best knowledge, SinGAN is the only well-known deep internal learning framework that performs general image manipulation. Therefore, SinIR and SinGAN are compared to provide meaningful comparisons in solving similar tasks and not for the purpose of replacing the latter. Additionally, recently introduced GAN-prior methods \cite{PanZDLLL20, GuSZ20} also tackle general image manipulation. However, they are not internal learning methods, and their outputs are obtained directly from the given images. (i.e., there is no separation between training and inference images like SinIR and SinGAN). This results in essential differences in methodology and solvable tasks.

Our main contributions can be summarized as follows.
\begin{itemize}
    \item To the best of our knowledge, SinIR is the first reconstruction-based deep internal learning framework for general image manipulation.
    
    \item By dint of a much simpler training objective (\emph{i.e.}, reconstruction), the training time of SinIR is vastly reduced compared to SinGAN \cite{Shaham19} that solves similar problems (trained \textbf{33.49 times faster} for \(500 \times 500\) images) as discussed in Section \ref{sec3.1}. This makes deep internal learning more plausible and practical.
    
    \item We show that random pixel shuffling enables successful manipulation of SinIR. Moreover, owing to this, SinIR obtains additional controllability over manipulation results. We give analyses with several tasks.
    
    \item As depicted in Figure \ref{main} and discussed in Section \ref{sec3.2} with quantitative evaluations, SinIR has competitive performance for various image manipulation tasks. Even though SinIR is trained with a single image, it produces visually pleasing results comparable to those of dedicated methods trained on large-scale datasets.

\end{itemize}

\section{Method}

\begin{figure*}[t]
  \centering
  \includegraphics[width=\linewidth]{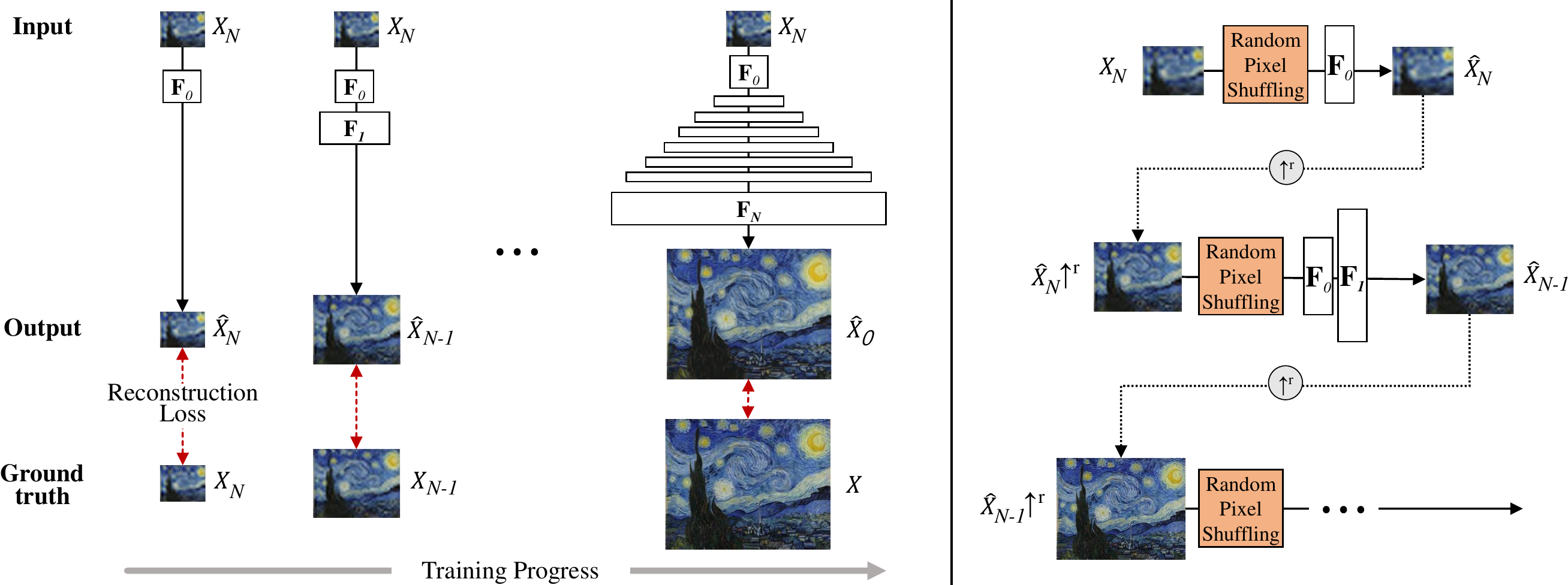}
  \vspace{-3mm}
  \caption{\textbf{Training of SinIR.}  We train a reconstruction network at each resolution level to refine images in a cascaded manner (Right). However, without random pixel shuffling, it is hard for SinIR to obtain meaningful results as explained in Section \ref{sec2.1.2} and Figure \ref{rps}. The outputs of each network are upsampled (\(\uparrow^r\)) and used as inputs at a one-level finer scale. SinIR is trained with progressively growing learning \cite{Karras18} (Left). Please see Section \ref{sec2.1} for details. Note that SinIR is fully-convolutional, and thus images of any size can be used for training and inference. The network details are in the supplement.}
  \label{network}
\end{figure*}

Our goal is, given \emph{one single natural image}, to train a model for general image manipulation in a much faster way. Although SinGAN \cite{Shaham19} solves similar problems, it suffers from prolonged training time with a complex GAN-based objective. Instead of using GANs, we start from a well-known unsupervised representation learning framework, autoencoders \cite{HintonSalakhutdinov2006b}, because of its \emph{reconstruction} objective that is much simpler than unconditional image generation.

However, the autoencoder was originally designed to be trained on large-scale datasets, and when trained on a single image, autoencoders may learn trivial identity mapping and reproduce meaningless samples if random inference images are provided for manipulation. Thus, for our model to learn better representations from a single image and achieve general image manipulation, we expand the capability of autoencoders with two methods: \textbf{cascaded multi-scale learning} and \textbf{random pixel shuffling}.

\subsection{\label{sec2.1}Training}

\subsubsection{Cascaded Multi-scale Learning}
A single natural image often contains various structures across different scales. To successfully learn these cross-scale visual properties, our model learns to \textbf{refine a downsampled training image to obtain the original image in a cascaded manner across multiple scales.} Our model consists of multiple networks that are responsible for refinement at each scale. Outputs of each network will be upsampled and fed into a network at a one-level finer scale (thus, inputs and outputs have the same resolution, but inputs are blurry). The networks at coarser scales will learn more to refine overall structures, while networks at finer scales will learn more to refine detailed textures.

Although there are subtle methodological differences, this multi-scale approach is a well-explored practice \cite{BurtA83, DentonCSF15, HuangLPHB17, ChenK17, ZhangXL17, LiFYWLY17, Wang0ZTKC18}. Especially to overcome the shortage of large-scale datasets, many deep internal learning frameworks such as ZSSR \cite{Shocher18zeroshot}, INGAN \cite{Shocher19} and SinGAN \cite{Shaham19} adopted this approach. Note that SinIR's architecture follows that of SinGAN in this work for simplicity

To train multi-scale networks, we opt for progressively growing learning \cite{Karras18}, as it is well-known as one of the most typical frameworks for multi-scale learning. Progressively growing multi-scale learning is widely used by many image synthesis methods \cite{Aigner18, Karras19, Zhang19,ChenK17,QiCJK18}, including SinGAN \cite{Shaham19}. They train multiple networks one by one from the coarsest scale while freezing previously trained networks. Such methods ease the difficulty of training by solving easier problems one by one.

\begin{figure*}[t]
  \centering
  \includegraphics[width=\linewidth]{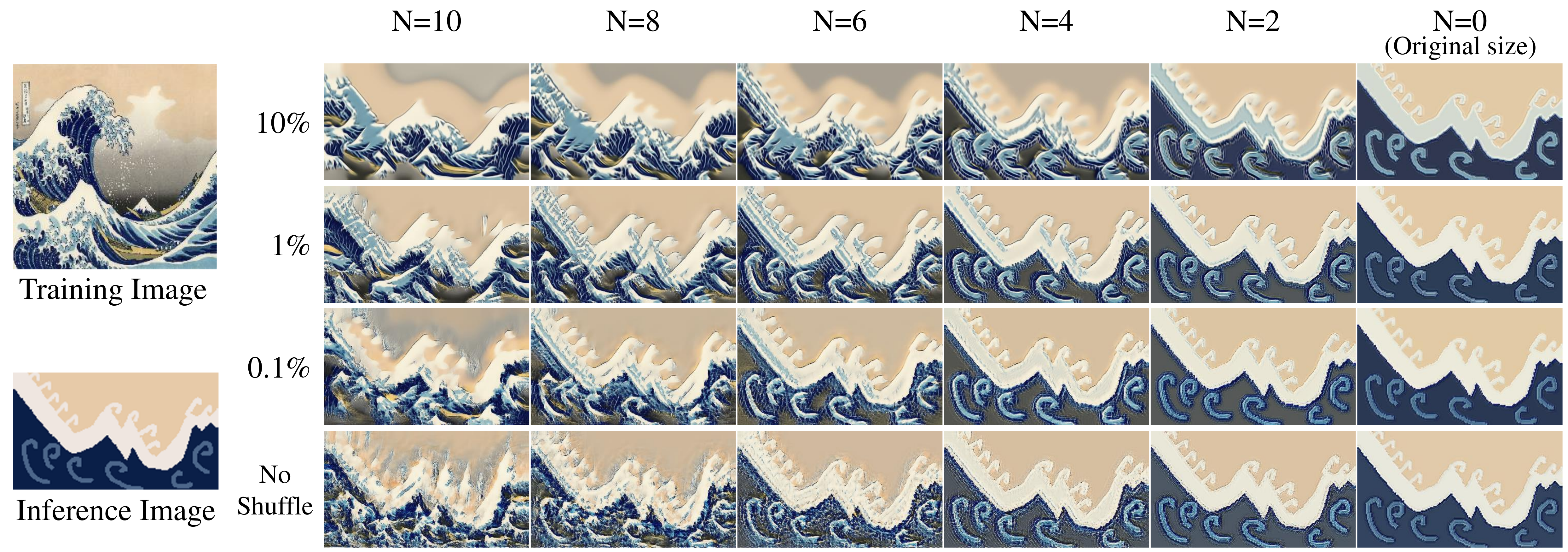}
  \vspace{-0.6cm}
  \caption{\label{rps}\textbf{Effects of percentages of randomly shuffled pixels and different inference starting scales.} Each row and column corresponds to different percentages and different inference starting scales. Also, note when we do not shuffle pixels of the training image, speckle-like or pecky stains are observed in manipulated inference images. This is not desirable as it hampers better reflection of original textures from the training image. On the other hand, when we randomly shuffle some pixels, such artifacts are effectively reduced. See Section \ref{sec2.1} and \ref{sec2.2} for details.}
\end{figure*}

Specifically, we downsample the training image and obtain ground-truth images for \(N+1\) scales, \(\{X_N, X_{N-1}, \dots, X_1\}\) and \(X_0 = X\), where \(X\) is the original training image. Then, at each \(n\)-th scale, we train \(n\)-th network (\(F_n\)) to reconstruct the one-level finer image. Thus, denoting \(\hat{X}_n = F_n(\hat{X}_{n+1}\uparrow^r\)) and \(\hat{X}_N = F_N(X_N)\) for the coarsest scale (\(\uparrow^r\) means upsampling by a factor \(r\)), our objective is
\begin{equation}
    \label{eq1}
    \min_{F_n} \mathcal{L}_{\text{rec}}(X_n, \hat{X}_n),
\end{equation}
where \(\mathcal{L}_{\text{rec}}\) is some reconstruction loss. Once we finish training one network, we freeze this network and add new network to be trained for a one-level finer scale.

\subsubsection{\label{sec2.1.2}Random Pixel Shuffling}
Although cascaded multi-scale learning gives a great chance to learn cross-scale representations, it is insufficient to achieve successful manipulation (see visual artifacts in the last row of Figure \ref{rps}). This is likely because of our training objective, reconstruction. As aforementioned, we use outputs of coarser scales as inputs of finer scales in a cascaded manner. Thus, if outputs of coarser scales show less diversity, it has the effect of showing limited variations in training samples to the next networks. Then it is likely that networks of finer scales will learn trivial or even identity mapping and fail to learn robust representations. In the case of GANs, this problem is naturally avoided as they use random noise (or latent code) for inputs. Besides, because of adversarial loss, they are less constrained in terms of outputs giving diverse samples, compared to the reconstruction loss. The effect of limited diversity is even more exacerbated when we use progressive growing of networks \cite{Karras18} for reconstruction because each network is trained only with \emph{fixed} outputs from \emph{frozen} networks. In other words, there is \emph{zero} diversity in inputs of each network.

However, this problem can be effectively mitigated with a simple technique inspired by denoising autoencoder \cite{Vincent08}: \textbf{random pixel shuffling}. An autoencoder may end up learning identity mapping and fail to learn robust representations. However, simply by imposing corruption to inputs, deterministic mapping of the autoencoder can be replaced with stochastic mapping, even giving the autoencoder a generative property, although it is trained with reconstruction loss. With this, the network can capture the main variation of the images, learning robust representations.

Noticing that the problem posed by denoising autoencoder is very analogous to ours, we apply a similar idea to SinIR. Instead of randomly setting some pixels to complete black as denoising autoencoder does, considering that we are using more complicated natural images, we randomly shuffle some pixels in the given single image so that the network can learn a more robust relationship between adjacent pixels. The effect of different types of corruption is explored in the supplement. But for simplicity, we use random pixel shuffling in this paper.

Rows of Figure \ref{rps} illustrates the effect of random pixel shuffling. When we do not use random pixel shuffling (last row), it shows severe artifacts that significantly harm manipulation quality. However, when we randomly shuffle \(0.05\%\) to \(5\%\) pixels, it shows much better manipulation quality. Interestingly, depending on the percentage, SinIR produces very different results. This behavior has various usages, as discussed in detail later. For example, this is utilized as a tool to control perception-distortion tradeoff \cite{Blau18} of super-resolution (Figure \ref{tradeoff}). Also, for artistic style transfer, this property can be used to control style-content tradeoff (Figure \ref{art}). It can also be used to adjust smoothness as it stands out in Figure \ref{rps}.

\subsubsection{Optimization}
We use the Adam optimizer (\(\beta_1=0.5, \beta_2=0.999\)) \cite{Kingma14} and the learning rate is 1e-4, unless mentioned otherwise. For the reconstruction loss, we use MSE (mean squared error) loss. However, it is well-known that MSE loss produces blurry images \cite{Zhao17}, thus we combine MSE loss and SSIM (structural similarity) loss \cite{Wang04}. \(\mathcal{L}_{\text{rec}}\) in Equation \ref{eq1} is
\begin{equation}
    \mathcal{L}_{\text{rec}} (A,B) = \mbox{MSE}(A, B) + (1 - \mbox{SSIM}(A, B)).
\end{equation}

\subsection{Inference: General Image Manipulation \label{sec2.2}}
The columns of Figure \ref{rps} depict the effect of different inference starting scales. After deciding on the starting scale, we properly downsample and forward inference images so that when it reaches the top scale (\(n=0\)), the size of the output becomes the same as the original. If the inference starting scale is close to the coarsest (\(n=N\)), the image is manipulated more globally and vice versa. Thus, depending on the inference image and the starting scale, we can achieve general image manipulation with a single trained model. For example, if we feed a clip-art to coarser scales, it performs the paint-to-image task. If we feed a classical painting to the same scales, it performs artistic style transfer. If we feed some images to the finest scale, it performs photo-realistic style transfer as color and tone are manipulated. A similar multi-scale inference scheme can be found in WCT \cite{LiFYWLY17} and SinGAN \cite{Shaham19}, which also employ multi-scale learning. 

\begin{table}
\centering
\setlength\tabcolsep{10pt}
\caption{\label{analysis}\textbf{Comparative analysis of training time.}  All results are obtained by averaging ten rounds on a single NVIDIA RTX 2080 Ti GPU.\\}
\begin{tabular}{lccc}\toprule
Image size  & \multirow{2}{*}{SinGAN} & SinIR & \multirow{2}{*}{Speedup} \\
(\# scales) & & (Ours) & \\
\midrule
\(125\)px (8)  & 36m 29s  & \textbf{1m 53s} & \(\times\)19.37 \\
\(250\)px (11) & 90m 39s  & \textbf{3m 39s} & \(\times\)24.84 \\
\(500\)px (13) & 267m 57s & \textbf{8m 0s}  & \(\times\)33.49 \\
\bottomrule
\end{tabular}
\end{table}

\section{Experiments\label{sec3}}
All experiments are conducted on the same machine with a single NVIDIA RTX 2080 Ti GPU. For a fair comparison, in the same way with SinGAN \cite{Shaham19}, we set scale factor \(r\) closely to 4/3, minimum and maximum dimension to 25px, 250px, and the number of scale \(N + 1\) is calculated from these parameters. Note that the maximum dimension constraint of SinGAN (250px) due to impractically long training time can be greatly relieved by SinIR's faster training speed (Table \ref{analysis}). Moreover, for the same reason, all results from SinIR in this section are obtained \textbf{in a few minutes}. In terms of iteration number, SinIR requires much fewer iterations as its training objective (\emph{i.e.}, reconstruction) is much simpler than unconditional image generation. Thus, SinIR is trained for 500 iterations at every scale, whereas SinGAN is trained for 6,000 iterations at every scale following the authors' best practice. For example, when we have 10 scales, SinGAN is trained for 60,000 iterations, and SinIR is trained for 5,000 iterations. As explained in Section \ref{sec1}, although comparisons are mainly conducted with SinGAN \cite{Shaham19}, this does not mean SinIR can completely replace SinGAN. The percentage of random pixel shuffling is set to 5e-4. For resampling, a bicubic kernel is used.

\subsection{Analysis\label{sec3.1}}

Table \ref{analysis} shows the training time of SinIR and SinGAN \cite{Shaham19}. SinIR is trained \textbf{24.84 times faster} compared to SinGAN, with \(250 \times 250\) images, which is the maximum dimension of training images presented in SinGAN paper. Moreover, SinIR is trained \textbf{33.49 times faster} with \(500 \times 500\) images. Owing to these results, SinIR makes deep internal learning for general image manipulation more plausible and practical.

\begin{figure*}[t]
  \centering
  \includegraphics[width=\linewidth]{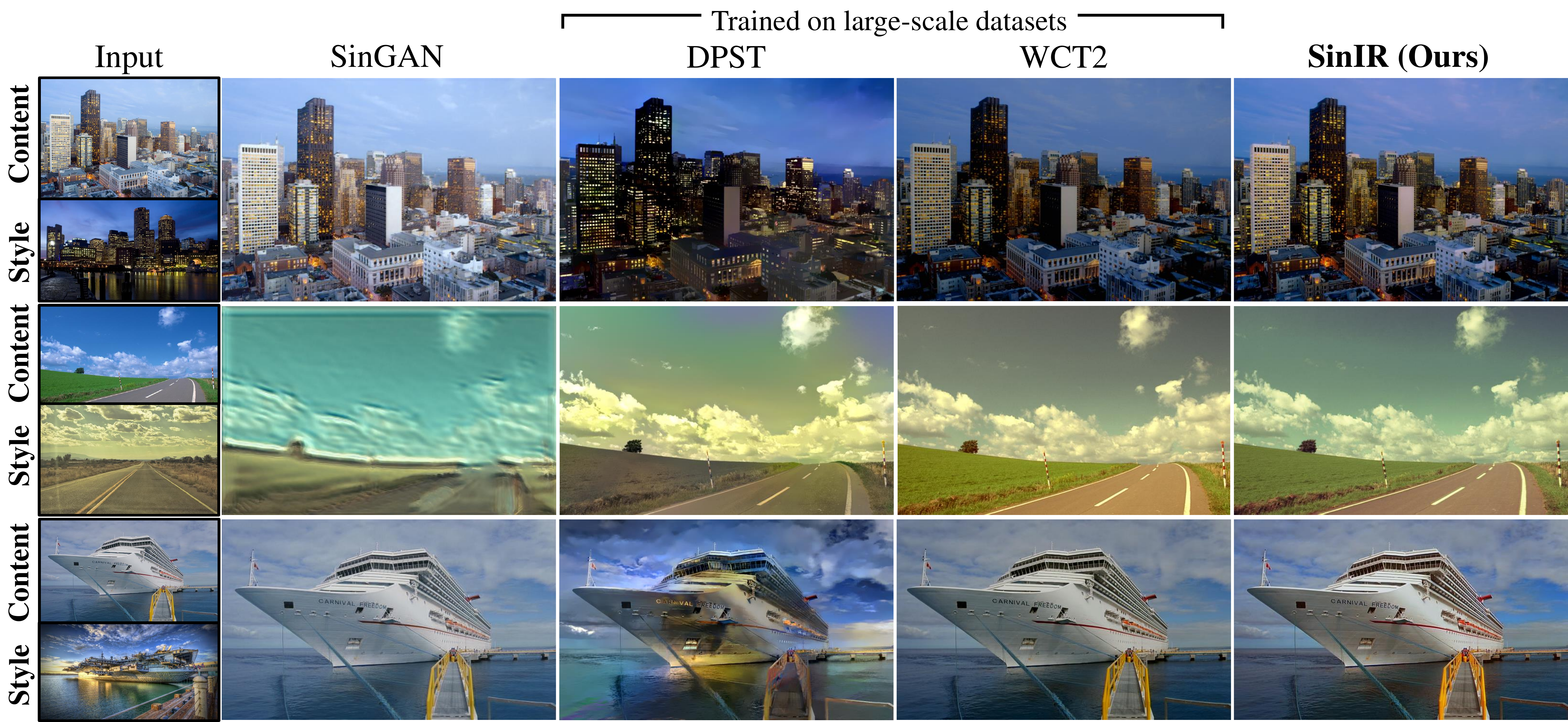}
  \vspace{-0.6cm}
  \caption{Visual results of \textbf{Photo-realistic Style Transfer}. \emph{DPST} and \emph{WCT2} are the results from \cite{Luan17} and \cite{Yoo19}.}
  \label{photo}
\end{figure*}

\begin{figure*}
  \centering
  \includegraphics[width=\linewidth]{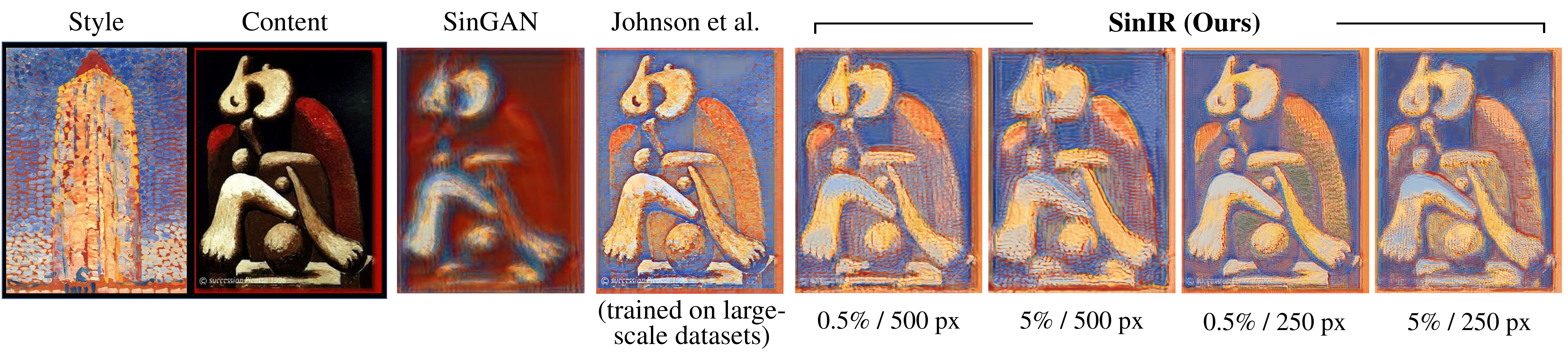}
  \vspace{-0.7cm}
  \caption{Visual results of \textbf{Artistic Style Transfer}. The numbers below the results of SinIR indicate the percentages of randomly shuffled pixels and the maximum dimension of the training image. These results show that SinIR has control over manipulation using random pixel shuffling. The inference starting scale for all SinIR results is \(n=N-4\). Note that Johnson et al. \cite{Johnson16} is a dedicated method trained externally. Best viewed when zoomed in.}
  \label{art}
\end{figure*}

\begin{figure*}
  \centering
  \includegraphics[width=\linewidth]{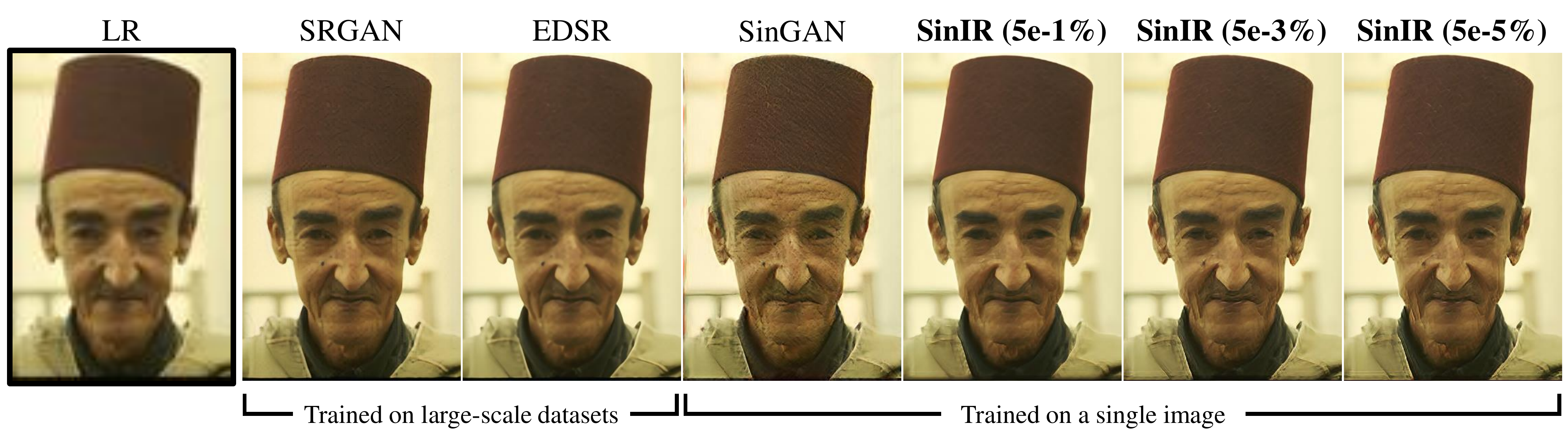}
  \vspace{-0.8cm}
  \caption{Visual results of \textbf{4X Super-resolution}. SRGAN \cite{Ledig17} and EDSR\cite{Lim17} are dedicated methods trained on large-scale datasets. See Section \ref{sr_section} for detailed analysis. Best viewed when zoomed in.}
  \label{pti}
  \vspace{-0.2cm}
\end{figure*}

\begin{table}
    \centering
    \caption{\label{ustable}\textbf{Results of the user study.} The numbers indicate preference rates. Photo ST and Artistic ST means photo-realistic and artistic style transfer.}
    \begin{tabular}{@{}lcccc@{}}\toprule
    \multicolumn{1}{c}{} & \multicolumn{1}{c}{SinIR (Ours)} & \multicolumn{1}{c}{SinGAN} & \multicolumn{1}{c}{Indecisive}\\
    \midrule
    Photo ST       & \textbf{90.00\%} & 6.75\%  & 3.25\%\\
    Artistic ST    & \textbf{76.95\%} & 18.70\% & 4.35\%\\
    Paint-to-image & \textbf{64.50\%} & 28.00\% & 8.50\%\\
    Editing        & \textbf{51.25\%} & 37.00\% & 11.75\%\\
    Harmonization  & \textbf{56.90\%} & 33.33\% & 9.76\%\\
    \bottomrule
    \end{tabular}
\end{table}

\subsection{Applications\label{sec3.2}}
We explore the capacity of SinIR for various image manipulation tasks. Inference images are not limited to 250px. All dedicated methods used for comparisons are not internal learning methods, thus require large-scale datasets. More results are included in the supplement. For application, a SinIR model is trained on an image with desired textures (\emph{e.g.}, style image for style-transfer), and then inference images (\emph{e.g.}, content image for style-transfer) are fed to this model for manipulation.

\paragraph{Photo-realistic Style Transfer.}
Photo-realistic style transfer demands strict preservation of the original content while transferring the color and tone of a given style image. To this end, we feed content images to finer scales so that the model does not manipulate overall structures except for the color and tone. In Figure \ref{photo}, we can observe that SinIR successfully transfers color and tone while not touching structures. On the other hand, SinGAN shows less capability for photo-realistic style transfer. For quantitative evaluation, we conducted a user study using randomly sampled images from a dedicated dataset provided by \cite{Luan17}. We showed 20 samples of SinIR and SinGAN to 20 subjects experienced in computer vision and asked to choose better samples. The preference rate of SinIR in Table \ref{ustable} was significantly higher than that of SinGAN with \textbf{a considerable margin of 83.25\%}, which aligns with the qualitative evaluation.

Note that the results of SinIR are similar to those of dedicated methods. Moreover, to further reduce training and inference time, we can optionally set scale factor \(r\) to 1 and the number of scales to 2 or so, as we do not need manipulation of overall structures. With this setting, we averaged 10 inferences of \(1024 \times 1024\) images. Impressively, SinIR took \textbf{less than a second} taking \textbf{0.189 s} and \textbf{0.381 s} for starting scales \(n=0\) and \(n=1\), while WCT2 took \textbf{2.030 s} and \textbf{4.799 s} with the fastest and slowest methods. Considering WCT2 was 830 times faster than previous methods, this result makes SinIR close to state-of-the-art, achieving \textbf{real-time photo-realistic style transfer}. However, the best scenarios for each method may be different. SinIR may be preferred when real-time inference with a fixed style is desirable and training samples are limited. In contrast, WCT2 may be preferred when training on large-scale datasets is feasible and transferring arbitrary styles is required.

\paragraph{Artistic Style Transfer.}

Artistic style transfer requires the successful blending of style and content. As Figure \ref{art} shows, SinIR is able to produce visually pleasing results, successfully blending the style and the content. Also, SinIR has additional controllability over manipulation results by adjusting the percentage of randomly shuffled pixels, as explained in Section \ref{sec2.1.2} and Figure \ref{rps}. In particular, when we increase the percentage, the given style is more aggressively blended, and vice versa (\(0.5\%\) vs. \(5\%\) in Figure \ref{art}). If users want to transfer larger textures, a larger maximum dimension of the training images may be preferred (250px vs. 500px in Figure \ref{art}. Please zoom and see the blob-like textures which became larger, for example). This option is less feasible with SinGAN \cite{Shaham19}, because it requires greatly prolonged training time with larger maximum dimension (\emph{e.g.}, 4.5 hours for \(500 \times 500\) images. See Section \ref{sec3.1} and Table \ref{analysis}). Also, SinGAN shows less capability on this task, obtaining heavily distorted images or limited difference from the content image. For quantitative evaluation, we created 20 samples from SinIR and SinGAN using images collected from the Web and showed them to 23 subjects experienced in computer vision. Aligning with our qualitative evaluation, the preference rate of SinIR was significantly higher than that of SinGAN \textbf{with a considerable margin of 58.26\%} (Table \ref{ustable}).

\begin{table}[t]
    \centering
    \caption{\label{srtable}\textbf{Results of 4X Super-resolution on the BSD100 benchmark.} The percentages of randomly shuffled pixels of SinIR in parentheses. For example, \emph{SinIR (5e-2)} means 0.05\(\%\) of pixels are shuffled during training.}
    \begin{tabular}{lcccc}\toprule
    \multicolumn{1}{c}{\multirow{2}{*}{}} & \multicolumn{1}{c}{{MS-}} & \multicolumn{1}{c}{\multirow{2}{*}{SSIM\(\uparrow\)}} & \multicolumn{1}{c}{\multirow{2}{*}{RMSE\(\downarrow\)}} & \multicolumn{1}{c}{\multirow{2}{*}{NIQE\(\downarrow\)}}\\
    \multicolumn{1}{c}{} & \multicolumn{1}{c}{SSIM\(\uparrow\)} & \multicolumn{1}{c}{} & \multicolumn{1}{c}{} & \multicolumn{1}{c}{}\\
    \midrule
    SRGAN & 0.933 & 0.640 & 16.33 & 3.407 \\
    EDSR  & \textbf{0.963} & \textbf{0.743} & \textbf{12.29} & 6.498 \\
    SinGAN         & 0.913 & 0.612 & 16.21 & 3.709 \\
    \midrule
    SinIR (5e-1)   & 0.915 & 0.635 & 16.28 & 4.155 \\
    SinIR (5e-2)   & 0.920 & 0.632 & 16.56 & 3.647 \\
    SinIR (5e-3)   & 0.918 & 0.623 & 17.14 & 3.410 \\
    SinIR (5e-4)   & 0.917 & 0.621 & 17.23 & \textbf{3.402} \\
    SinIR (5e-5)   & 0.916 & 0.619 & 17.34 & 3.409 \\
    \bottomrule
    \end{tabular}
\vspace{-0.5cm}
\end{table}

\begin{figure*}[t]
  \centering
  \includegraphics[width=\linewidth]{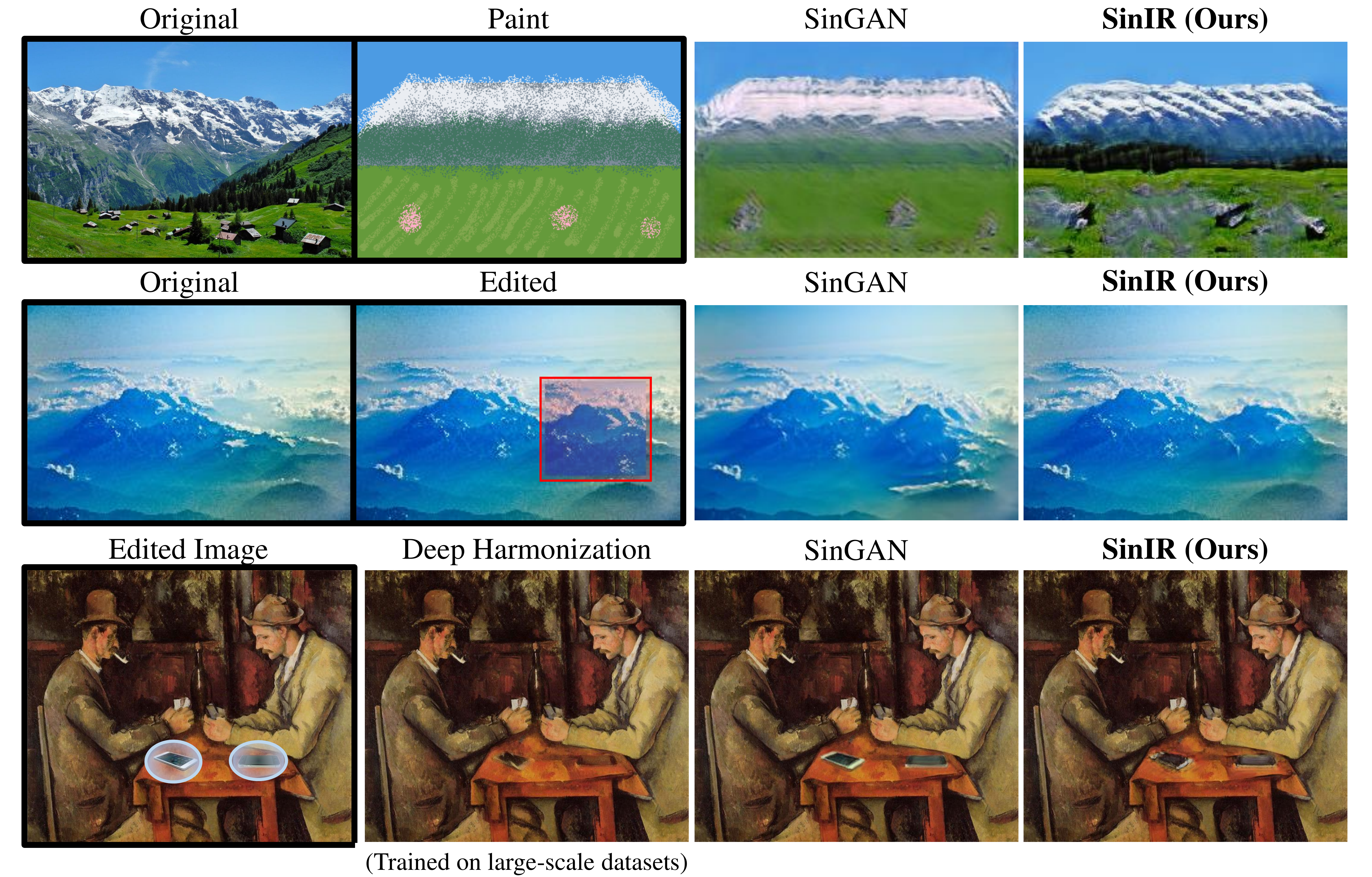}
  \vspace{-0.8cm}
  \caption{Visual results of \textbf{Paint-to-image} (Top), \textbf{Editing} (Middle. Edited region in a square), and \textbf{Harmonization} (Bottom. Pasted objects in circles). \emph{Deep Harmonization} indicates Deep Paint Harmonization \cite{Luan18}, a dedicated external method.}
  \label{peh}
\end{figure*}

\begin{figure}[h]
  \centering
  \includegraphics[width=\linewidth]{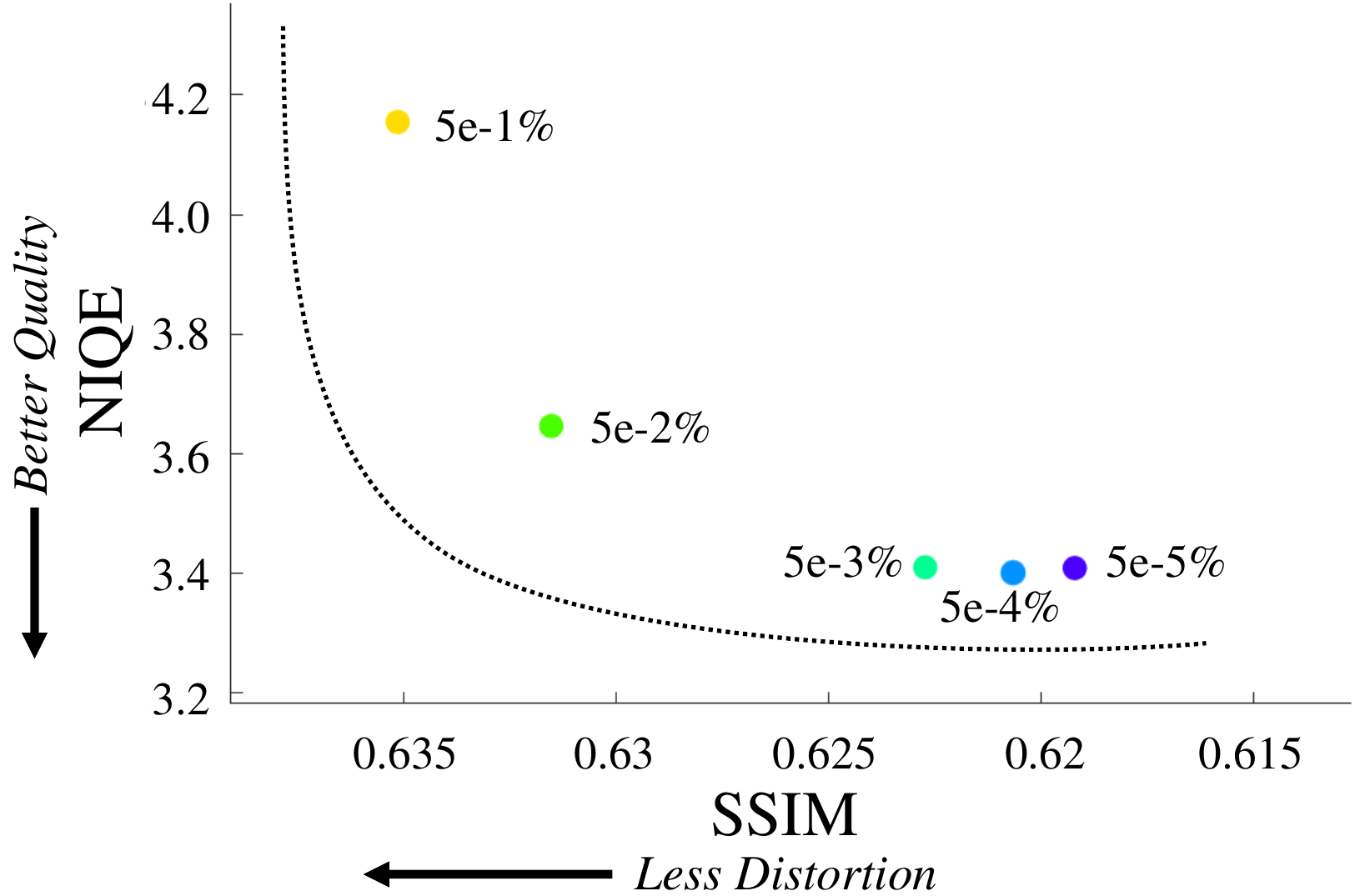}
  \vspace{-0.6cm}
  \caption{\textbf{Controllability over perception-distortion trade-off} \cite{Blau18} \textbf{of SinIR.} As the percentage of randomly shuffled pixels increases, it shows less distortion but worse perceptual quality, vice versa. The numbers indicate the percentage of randomly shuffled pixels during training.}
  \label{tradeoff}
  \vspace{-0.4cm}
\end{figure}

\paragraph{\label{sr_section}Super-resolution.}
For super-resolution, we need to refine detailed textures, but overall structures should largely remain unchanged as it would otherwise lead to unnecessary distortion and poor quality. For this reason, SinIR generally shows better performance with fewer scales. For example, we train SinIR using 2 scales with a scale factor of 2. Instead, we set the number of kernels to 256, the iteration number at each scale to 1000 (2000 iterations in total), and the learning rate to 0.001 so that the overall model can still be trained sufficiently with enough capacity. Also, anti-aliasing is recommended when downsampling the training image to suppress undesirable artifacts that may lead to critical degradation in case of super-resolution. During inference, similarly to SinGAN \cite{Shaham19}, the last network is used for all scales as it better handles detailed textures. The LR image is properly upsampled before it is fed to the model to obtain the final HR image with the target size.

Table \ref{srtable} shows 4X super-resolution scores using BSD100 dataset \cite{Martin01}, in terms of distortion (RMSE, SSIM, and MS-SSIM) and perception (NIQE) quality. These are two conflicting criteria for super-resolution often found in a trade-off relationship \cite{Blau18}. In general, SinIR shows better perceptual scores that are similar to or even better than SRGAN \cite{Ledig17} that is externally trained on a large-scale dataset. Compared to SinGAN \cite{Shaham19} that is also trained on a single image, SinIR shows better perceptual quality. In terms of distortion score, SinIR was slightly better in MS-SSIM and SSIM and slightly inferior in RMSE than SinGAN.

Interestingly, by adjusting the percentage of randomly shuffled pixels during training (Section \ref{sec2.1.2}), the perception-distortion trade-off can be controlled to some extent by SinIR. When we increase the percentage, the distortion score improves while the perception score becomes worse, and vice versa, as depicted in Figure \ref{tradeoff}.

\paragraph{Paint-to-Image.}
Paint-to-image aims to obtain natural-looking images from simple drawings. As shown in the first row of Figure \ref{peh}, SinIR produces naturally textured images using the visual characteristics of the training image. For quantitative evaluation, we collected images from the Web and created simple drawings to obtain 20 samples from SinIR and SinGAN \cite{Shaham19}. Then we showed them to 20 subjects experienced in computer vision and asked them to choose better results. SinIR was preferred to SinGAN \textbf{with a margin of 36\%} as shown in Table \ref{ustable}.

\paragraph{Editing.}
For this task, we aim to obtain natural-looking images from edited images by blending an edited region with adjacent parts. We first manipulate edited images from SinIR, then combine them with the original using masks like SinGAN \cite{Shaham19}. The second row of Figure \ref{peh} shows that SinIR obtains visually consistent results. For quantitative evaluation, we created 20 edited images from the images collected from the Web. Then we showed the results of SinIR and SinGAN to 20 subjects, and SinIR was preferred to SinGAN \textbf{with a margin of 14.25\%}.

\paragraph{Harmonization.}
Image harmonization aims at the natural blending of pasted alien objects and original images. We manipulate edited images and combine them with the original using masks like SinGAN \cite{Shaham19}. The last row of Figure \ref{peh} shows that the results of SinIR are comparable to those of SinGAN and Deep Paint Harmonization \cite{Luan18}, a dedicated method. We conducted a user study using randomly sampled images from a dedicated dataset provided by \cite{Luan18}. We showed 20 samples to 21 subjects experienced in computer vision. As Table \ref{ustable} shows, SinIR was preferred to SinGAN \textbf{with a margin of 23.57\%}.

\section{Discussion}
We introduce SinIR, an internal learning framework trained on a single image for general image manipulation. SinIR learns reconstruction instead of utilizing the adversarial loss in SinGAN \cite{Shaham19}. Owing to this, SinIR is trained much faster (Table \ref{analysis}). However, to obtain visually pleasing manipulation results, we need to train SinIR with cascaded multi-scale learning and random pixel shuffling. Especially, random pixel shuffling is necessary as our training objective is reconstruction. Achieving general image manipulation with reduced training time, SinIR makes the real-world application of deep internal learning more practical with faster training speed and better visual results.

For future work, although we used progressively growing learning \cite{Karras18} for multi-scale learning as it is a well-known technique, more dedicated methods can be explored. Also, while we use only one training image in this work, the effect of training with multiple images can be further explored. Effective methods to handle extreme cases where there are fewer internal references in training images can be explored. Lastly, considering recent findings on memorization and generalization of neural networks with experiments with a single image \cite{ZhangBHMS20}, we find that sophisticated corruption techniques as in \cite{YuBSAL19} and \cite{KrullBJ19} can be considered instead of simple random pixel shuffling for more robust texture transfer.

\bibliography{mybib}

\begin{thebibliography}{58}
\providecommand{\natexlab}[1]{#1}
\providecommand{\url}[1]{\texttt{#1}}
\expandafter\ifx\csname urlstyle\endcsname\relax
  \providecommand{\doi}[1]{doi: #1}\else
  \providecommand{\doi}{doi: \begingroup \urlstyle{rm}\Url}\fi

\bibitem[Aigner \& K{\"{o}}rner(2018)Aigner and K{\"{o}}rner]{Aigner18}
Aigner, S. and K{\"{o}}rner, M.
\newblock Futuregan: Anticipating the future frames of video sequences using
  spatio-temporal 3d convolutions in progressively growing autoencoder gans.
\newblock \emph{arXiv: 1810.01325}, 2018.

\bibitem[Bagon et~al.(2008)Bagon, Boiman, and Irani]{Bagon08}
Bagon, S., Boiman, O., and Irani, M.
\newblock What is a good image segment? {A} unified approach to segment
  extraction.
\newblock In \emph{ECCV}, 2008.

\bibitem[Bahat \& Irani(2016)Bahat and Irani]{Bahat16}
Bahat, Y. and Irani, M.
\newblock Blind dehazing using internal patch recurrence.
\newblock In \emph{ICCP}, 2016.

\bibitem[Bell{-}Kligler et~al.(2019)Bell{-}Kligler, Shocher, and
  Irani]{Bell-Kligler19}
Bell{-}Kligler, S., Shocher, A., and Irani, M.
\newblock Blind super-resolution kernel estimation using an internal-gan.
\newblock In \emph{NeurIPS}, 2019.

\bibitem[Bengio et~al.(2006)Bengio, Lamblin, Popovici, and
  Larochelle]{Bengio06}
Bengio, Y., Lamblin, P., Popovici, D., and Larochelle, H.
\newblock Greedy layer-wise training of deep networks.
\newblock In \emph{NeurIPS}, 2006.

\bibitem[Bergmann et~al.(2017)Bergmann, Jetchev, and Vollgraf]{Bergmann17}
Bergmann, U., Jetchev, N., and Vollgraf, R.
\newblock Learning texture manifolds with the periodic spatial {GAN}.
\newblock In \emph{ICML}, 2017.

\bibitem[Blau \& Michaeli(2018)Blau and Michaeli]{Blau18}
Blau, Y. and Michaeli, T.
\newblock The perception-distortion tradeoff.
\newblock In \emph{CVPR}, 2018.

\bibitem[Burt \& Adelson(1983)Burt and Adelson]{BurtA83}
Burt, P.~J. and Adelson, E.~H.
\newblock The laplacian pyramid as a compact image code.
\newblock \emph{{IEEE} Trans. Commun.}, 1983.

\bibitem[Chen \& Koltun(2017)Chen and Koltun]{ChenK17}
Chen, Q. and Koltun, V.
\newblock Photographic image synthesis with cascaded refinement networks.
\newblock In \emph{ICCV}, 2017.

\bibitem[Denton et~al.(2015)Denton, Chintala, Szlam, and Fergus]{DentonCSF15}
Denton, E.~L., Chintala, S., Szlam, A., and Fergus, R.
\newblock Deep generative image models using a laplacian pyramid of adversarial
  networks.
\newblock In \emph{NeurIPS}, 2015.

\bibitem[Efros \& Leung(1999)Efros and Leung]{Efros99}
Efros, A.~A. and Leung, T.~K.
\newblock Texture synthesis by non-parametric sampling.
\newblock In \emph{ICCV}, 1999.

\bibitem[Fan et~al.(2019)Fan, Yin, Chen, Wang, Avil{\'{e}}s{-}Rivero, Li,
  Sch{\"{o}}nlieb, Lischinski, and Chen]{Fan19}
Fan, Q., Yin, Y., Chen, D., Wang, Y., Avil{\'{e}}s{-}Rivero, A.~I., Li, R.,
  Sch{\"{o}}nlieb, C., Lischinski, D., and Chen, B.
\newblock Deep reflection prior.
\newblock \emph{arXiv: 1912.03623}, 2019.

\bibitem[Gandelsman et~al.(2019)Gandelsman, Shocher, and Irani]{Gandelsman19}
Gandelsman, Y., Shocher, A., and Irani, M.
\newblock "double-dip": Unsupervised image decomposition via coupled
  deep-image-priors.
\newblock In \emph{CVPR}, 2019.

\bibitem[Glasner et~al.(2009)Glasner, Bagon, and Irani]{Glasner09}
Glasner, D., Bagon, S., and Irani, M.
\newblock Super-resolution from a single image.
\newblock In \emph{ICCV}, 2009.

\bibitem[Goodfellow et~al.(2014)Goodfellow, Pouget{-}Abadie, Mirza, Xu,
  Warde{-}Farley, Ozair, Courville, and Bengio]{Goodfellow14}
Goodfellow, I.~J., Pouget{-}Abadie, J., Mirza, M., Xu, B., Warde{-}Farley, D.,
  Ozair, S., Courville, A.~C., and Bengio, Y.
\newblock Generative adversarial nets.
\newblock In \emph{NeurIPS}, 2014.

\bibitem[Gu et~al.(2020)Gu, Shen, and Zhou]{GuSZ20}
Gu, J., Shen, Y., and Zhou, B.
\newblock Image processing using multi-code {GAN} prior.
\newblock In \emph{CVPR}, 2020.

\bibitem[Gulrajani et~al.(2017)Gulrajani, Ahmed, Arjovsky, Dumoulin, and
  Courville]{Gulrajani17}
Gulrajani, I., Ahmed, F., Arjovsky, M., Dumoulin, V., and Courville, A.~C.
\newblock Improved training of wasserstein gans.
\newblock In \emph{NeurIPS}, 2017.

\bibitem[He \& Sun(2012)He and Sun]{He12}
He, K. and Sun, J.
\newblock Statistics of patch offsets for image completion.
\newblock In \emph{ECCV}, 2012.

\bibitem[He et~al.(2016)He, Zhang, Ren, and Sun]{He16}
He, K., Zhang, X., Ren, S., and Sun, J.
\newblock Deep residual learning for image recognition.
\newblock In \emph{CVPR}, 2016.

\bibitem[Hinton \& Salakhutdinov(2006)Hinton and
  Salakhutdinov]{HintonSalakhutdinov2006b}
Hinton, G.~E. and Salakhutdinov, R.~R.
\newblock Reducing the dimensionality of data with neural networks.
\newblock \emph{Science}, 313:\penalty0 504--507, 2006.

\bibitem[Huang et~al.(2017{\natexlab{a}})Huang, Liu, van~der Maaten, and
  Weinberger]{Huang17}
Huang, G., Liu, Z., van~der Maaten, L., and Weinberger, K.~Q.
\newblock Densely connected convolutional networks.
\newblock In \emph{CVPR}, 2017{\natexlab{a}}.

\bibitem[Huang et~al.(2017{\natexlab{b}})Huang, Li, Poursaeed, Hopcroft, and
  Belongie]{HuangLPHB17}
Huang, X., Li, Y., Poursaeed, O., Hopcroft, J.~E., and Belongie, S.~J.
\newblock Stacked generative adversarial networks.
\newblock In \emph{CVPR}, 2017{\natexlab{b}}.

\bibitem[Jetchev et~al.(2016)Jetchev, Bergmann, and Vollgraf]{Jetchev16}
Jetchev, N., Bergmann, U., and Vollgraf, R.
\newblock Texture synthesis with spatial generative adversarial networks.
\newblock \emph{arXiv: 1611.08207}, 2016.

\bibitem[Johnson et~al.(2016)Johnson, Alahi, and Fei{-}Fei]{Johnson16}
Johnson, J., Alahi, A., and Fei{-}Fei, L.
\newblock Perceptual losses for real-time style transfer and super-resolution.
\newblock In \emph{ECCV}, 2016.

\bibitem[Karras et~al.(2018)Karras, Aila, Laine, and Lehtinen]{Karras18}
Karras, T., Aila, T., Laine, S., and Lehtinen, J.
\newblock Progressive growing of gans for improved quality, stability, and
  variation.
\newblock In \emph{ICLR}, 2018.

\bibitem[Karras et~al.(2019)Karras, Laine, and Aila]{Karras19}
Karras, T., Laine, S., and Aila, T.
\newblock A style-based generator architecture for generative adversarial
  networks.
\newblock In \emph{CVPR}, 2019.

\bibitem[Kingma \& Ba(2015)Kingma and Ba]{Kingma14}
Kingma, D.~P. and Ba, J.
\newblock Adam: {A} method for stochastic optimization.
\newblock In \emph{ICLR}, 2015.

\bibitem[Krull et~al.(2019)Krull, Buchholz, and Jug]{KrullBJ19}
Krull, A., Buchholz, T., and Jug, F.
\newblock Noise2void - learning denoising from single noisy images.
\newblock In \emph{CVPR}, 2019.

\bibitem[Lecun et~al.(1998)Lecun, Bottou, Bengio, and Haffner]{Lecun98}
Lecun, Y., Bottou, L., Bengio, Y., and Haffner, P.
\newblock Gradient-based learning applied to document recognition.
\newblock In \emph{Proceedings of the IEEE}, 1998.

\bibitem[Ledig et~al.(2017)Ledig, Theis, Huszar, Caballero, Cunningham, Acosta,
  Aitken, Tejani, Totz, Wang, and Shi]{Ledig17}
Ledig, C., Theis, L., Huszar, F., Caballero, J., Cunningham, A., Acosta, A.,
  Aitken, A.~P., Tejani, A., Totz, J., Wang, Z., and Shi, W.
\newblock Photo-realistic single image super-resolution using a generative
  adversarial network.
\newblock In \emph{CVPR}, 2017.

\bibitem[Li \& Wand(2016)Li and Wand]{Li16}
Li, C. and Wand, M.
\newblock Precomputed real-time texture synthesis with markovian generative
  adversarial networks.
\newblock In \emph{ECCV}, 2016.

\bibitem[Li et~al.(2017)Li, Fang, Yang, Wang, Lu, and Yang]{LiFYWLY17}
Li, Y., Fang, C., Yang, J., Wang, Z., Lu, X., and Yang, M.
\newblock Universal style transfer via feature transforms.
\newblock In \emph{NIPS}, 2017.

\bibitem[Lim et~al.(2017)Lim, Son, Kim, Nah, and Lee]{Lim17}
Lim, B., Son, S., Kim, H., Nah, S., and Lee, K.~M.
\newblock Enhanced deep residual networks for single image super-resolution.
\newblock In \emph{CVPR}, 2017.

\bibitem[Luan et~al.(2017)Luan, Paris, Shechtman, and Bala]{Luan17}
Luan, F., Paris, S., Shechtman, E., and Bala, K.
\newblock Deep photo style transfer.
\newblock In \emph{CVPR}, 2017.

\bibitem[Luan et~al.(2018)Luan, Paris, Shechtman, and Bala]{Luan18}
Luan, F., Paris, S., Shechtman, E., and Bala, K.
\newblock Deep painterly harmonization.
\newblock \emph{Comput. Graph. Forum}, 37\penalty0 (4):\penalty0 95--106, 2018.

\bibitem[Maas et~al.(2013)Maas, Hannun, and Ng]{Maas13}
Maas, A.~L., Hannun, A.~Y., and Ng, A.~Y.
\newblock Rectifier nonlinearities improve neural network acoustic models.
\newblock In \emph{ICML Workshop on Deep Learning for Audio, Speech and
  Language Processing}, 2013.

\bibitem[Martin et~al.(2001)Martin, Fowlkes, Tal, and Malik]{Martin01}
Martin, D., Fowlkes, C., Tal, D., and Malik, J.
\newblock A database of human segmented natural images and its application to
  evaluating segmentation algorithms and measuring ecological statistics.
\newblock In \emph{ICCV}, 2001.

\bibitem[Mastan \& Raman(2020)Mastan and Raman]{MastanR20}
Mastan, I.~D. and Raman, S.
\newblock {DCIL:} deep contextual internal learning for image restoration and
  image retargeting.
\newblock In \emph{WACV}, 2020.

\bibitem[Pan et~al.(2020)Pan, Zhan, Dai, Lin, Loy, and Luo]{PanZDLLL20}
Pan, X., Zhan, X., Dai, B., Lin, D., Loy, C.~C., and Luo, P.
\newblock Exploiting deep generative prior for versatile image restoration and
  manipulation.
\newblock In \emph{ECCV}, 2020.

\bibitem[Qi et~al.(2018)Qi, Chen, Jia, and Koltun]{QiCJK18}
Qi, X., Chen, Q., Jia, J., and Koltun, V.
\newblock Semi-parametric image synthesis.
\newblock In \emph{CVPR}, 2018.

\bibitem[Ren et~al.(2019)Ren, Zhang, Wang, Hu, and Zuo]{Ren19}
Ren, D., Zhang, K., Wang, Q., Hu, Q., and Zuo, W.
\newblock Neural blind deconvolution using deep priors.
\newblock \emph{arXiv: 1908.02197}, 2019.

\bibitem[Shaham et~al.(2019)Shaham, Dekel, and Michaeli]{Shaham19}
Shaham, T.~R., Dekel, T., and Michaeli, T.
\newblock Singan: Learning a generative model from a single natural image.
\newblock In \emph{ICCV}, 2019.

\bibitem[Shocher et~al.(2018)Shocher, Cohen, and Irani]{Shocher18zeroshot}
Shocher, A., Cohen, N., and Irani, M.
\newblock "zero-shot" super-resolution using deep internal learning.
\newblock In \emph{CVPR}, 2018.

\bibitem[Shocher et~al.(2019)Shocher, Bagon, Isola, and Irani]{Shocher19}
Shocher, A., Bagon, S., Isola, P., and Irani, M.
\newblock Ingan: Capturing and retargeting the "dna" of a natural image.
\newblock In \emph{ICCV}, 2019.

\bibitem[Ulyanov et~al.(2016)Ulyanov, Vedaldi, and Lempitsky]{Ulyanov16}
Ulyanov, D., Vedaldi, A., and Lempitsky, V.~S.
\newblock Instance normalization: The missing ingredient for fast stylization.
\newblock \emph{arXive: 1607.08022}, 2016.

\bibitem[Ulyanov et~al.(2018)Ulyanov, Vedaldi, and Lempitsky]{Ulyanov18}
Ulyanov, D., Vedaldi, A., and Lempitsky, V.~S.
\newblock Deep image prior.
\newblock In \emph{CVPR}, 2018.

\bibitem[Vincent et~al.(2008)Vincent, Larochelle, Bengio, and
  Manzagol]{Vincent08}
Vincent, P., Larochelle, H., Bengio, Y., and Manzagol, P.
\newblock Extracting and composing robust features with denoising autoencoders.
\newblock In \emph{ICML}, 2008.

\bibitem[Wang et~al.(2018)Wang, Liu, Zhu, Tao, Kautz, and
  Catanzaro]{Wang0ZTKC18}
Wang, T., Liu, M., Zhu, J., Tao, A., Kautz, J., and Catanzaro, B.
\newblock High-resolution image synthesis and semantic manipulation with
  conditional gans.
\newblock In \emph{CVPR}, 2018.

\bibitem[Wang et~al.(2004)Wang, Bovik, Sheikh, and Simoncelli]{Wang04}
Wang, Z., Bovik, A.~C., Sheikh, H.~R., and Simoncelli, E.~P.
\newblock Image quality assessment: from error visibility to structural
  similarity.
\newblock \emph{{IEEE} Trans. Image Process.}, 13\penalty0 (4):\penalty0
  600--612, 2004.

\bibitem[Yoo et~al.(2019)Yoo, Uh, Chun, Kang, and Ha]{Yoo19}
Yoo, J., Uh, Y., Chun, S., Kang, B., and Ha, J.
\newblock Photorealistic style transfer via wavelet transforms.
\newblock In \emph{ICCV}, 2019.

\bibitem[Yu et~al.(2019)Yu, Barnes, Shechtman, Amirghodsi, and
  Luk{\'{a}}c]{YuBSAL19}
Yu, N., Barnes, C., Shechtman, E., Amirghodsi, S., and Luk{\'{a}}c, M.
\newblock Texture mixer: {A} network for controllable synthesis and
  interpolation of texture.
\newblock In \emph{CVPR}, 2019.

\bibitem[Zhang et~al.(2020)Zhang, Bengio, Hardt, Mozer, and
  Singer]{ZhangBHMS20}
Zhang, C., Bengio, S., Hardt, M., Mozer, M.~C., and Singer, Y.
\newblock Identity crisis: Memorization and generalization under extreme
  overparameterization.
\newblock In \emph{ICLR}, 2020.

\bibitem[Zhang et~al.(2017)Zhang, Xu, and Li]{ZhangXL17}
Zhang, H., Xu, T., and Li, H.
\newblock Stackgan: Text to photo-realistic image synthesis with stacked
  generative adversarial networks.
\newblock In \emph{ICCV}, 2017.
\newblock URL \url{https://doi.org/10.1109/ICCV.2017.629}.

\bibitem[Zhang et~al.(2019{\natexlab{a}})Zhang, Mai, Jin, Wang, Xu, and
  Collomosse]{Zhang19vidinpaint}
Zhang, H., Mai, L., Jin, H., Wang, Z., Xu, N., and Collomosse, J.~P.
\newblock An internal learning approach to video inpainting.
\newblock In \emph{ICCV}, 2019{\natexlab{a}}.

\bibitem[Zhang et~al.(2019{\natexlab{b}})Zhang, Zhang, Liu, Shen, Zhang, and
  Zhao]{Zhang19restor}
Zhang, L., Zhang, L., Liu, X., Shen, Y., Zhang, S., and Zhao, S.
\newblock Zero-shot restoration of back-lit images using deep internal
  learning.
\newblock In \emph{ACM Multimedia}, 2019{\natexlab{b}}.

\bibitem[Zhang et~al.(2019{\natexlab{c}})Zhang, Li, and Yu]{Zhang19}
Zhang, Z., Li, M., and Yu, J.
\newblock {D2PGGAN:} two discriminators used in progressive growing of {GANS}.
\newblock In \emph{ICASSP}, 2019{\natexlab{c}}.

\bibitem[Zhao et~al.(2017)Zhao, Gallo, Frosio, and Kautz]{Zhao17}
Zhao, H., Gallo, O., Frosio, I., and Kautz, J.
\newblock Loss functions for image restoration with neural networks.
\newblock \emph{{IEEE} Trans. Computational Imaging}, 3\penalty0 (1):\penalty0
  47--57, 2017.

\bibitem[Zhou et~al.(2018)Zhou, Zhu, Bai, Lischinski, Cohen{-}Or, and
  Huang]{Zhou018}
Zhou, Y., Zhu, Z., Bai, X., Lischinski, D., Cohen{-}Or, D., and Huang, H.
\newblock Non-stationary texture synthesis by adversarial expansion.
\newblock \emph{{ACM} Trans. Graph.}, 37\penalty0 (4):\penalty0 49:1--49:13,
  2018.

\end{thebibliography}
\bibliographystyle{icml2021}

\clearpage
\begin{appendices}

\begin{figure}[!ht]
  \setlength{\linewidth}{\textwidth}
  \setlength{\hsize}{\textwidth}
  \centering
  \includegraphics[width=\textwidth]{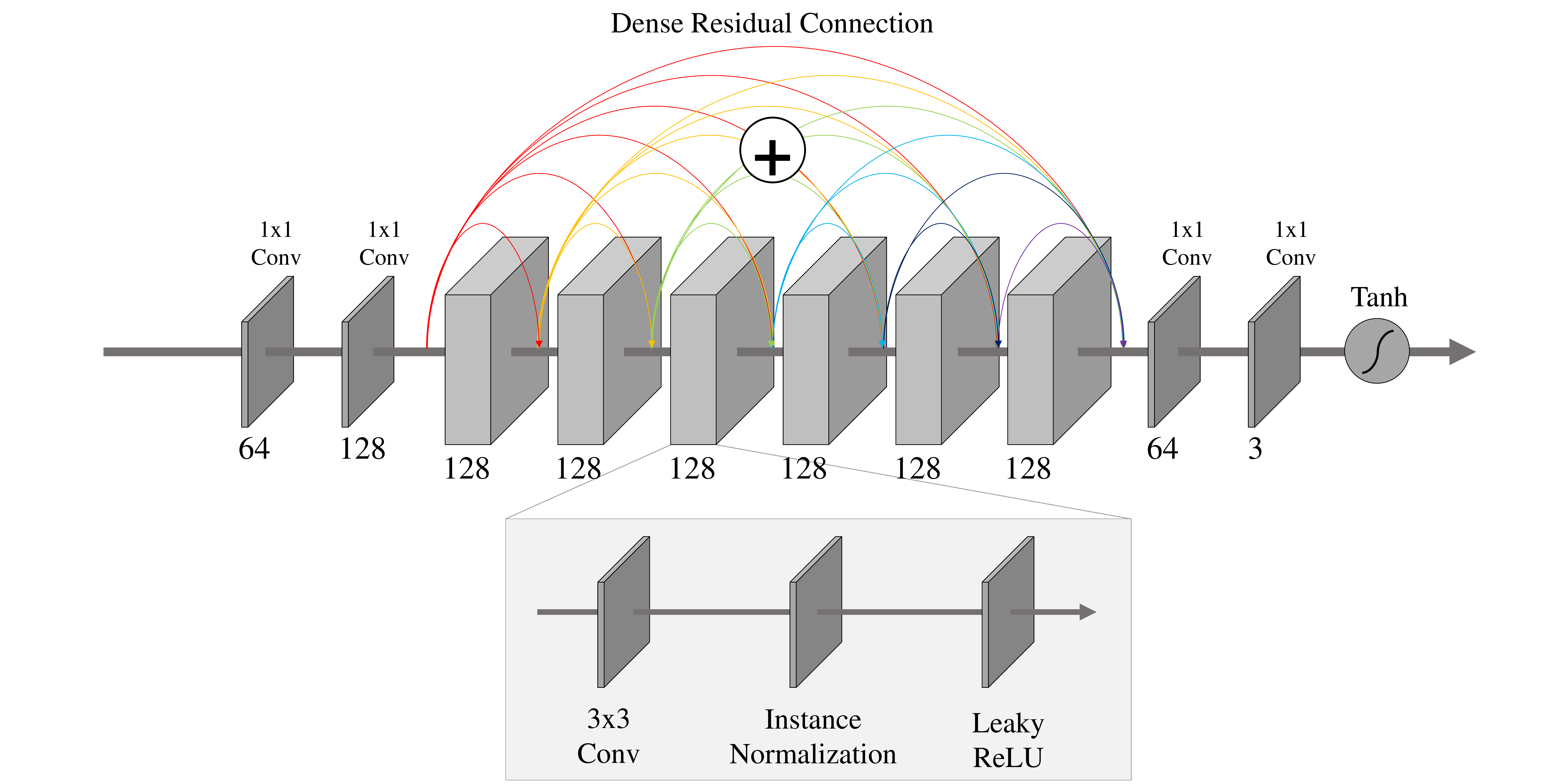}
  \caption{\textbf{Network Architecture of SinIR.} The numbers below each layer indicate the number of convolutional kernels. Here we assume that the input and the output are RGB images.}
  \label{net_archi}
\end{figure}

\section{Network Architecture}
All the networks at every scale use the same architecture described in Figure \ref{net_archi}. Each network consists of two \(1 \times 1\) convolutional layers which map RGB images to feature space, six convolutional blocks which are densely connected \cite{Huang17} with residual operation \cite{He16} (not concatenation), and two \(1 \times 1\) convolutional layers which render features to RGB images. Each convolutional block has one \(3 \times 3\) convolutional layer, an instance normalization layer \cite{Ulyanov16} and a LeakyReLU activation layer (negative slope = 0.2) \cite{Maas13}. We do not use pooling or unpooling inside a network, and thus the inputs and the outputs of each network have the same spatial dimension. Reflection padding is used before \(3 \times 3\) convolutional layer. Tanh function is used to obtain the final output.

\section{Additional Results}
We show more results for all tasks presented in the original paper. The setting described in the paper is used for all results. While SinIR and SinGAN \cite{Shaham19} are internal methods trained on a single image, all dedicated methods used for comparisons are external methods trained on large-scale datasets.

\begin{figure*}[!t]
    \centering
    \includegraphics[width=\textwidth]{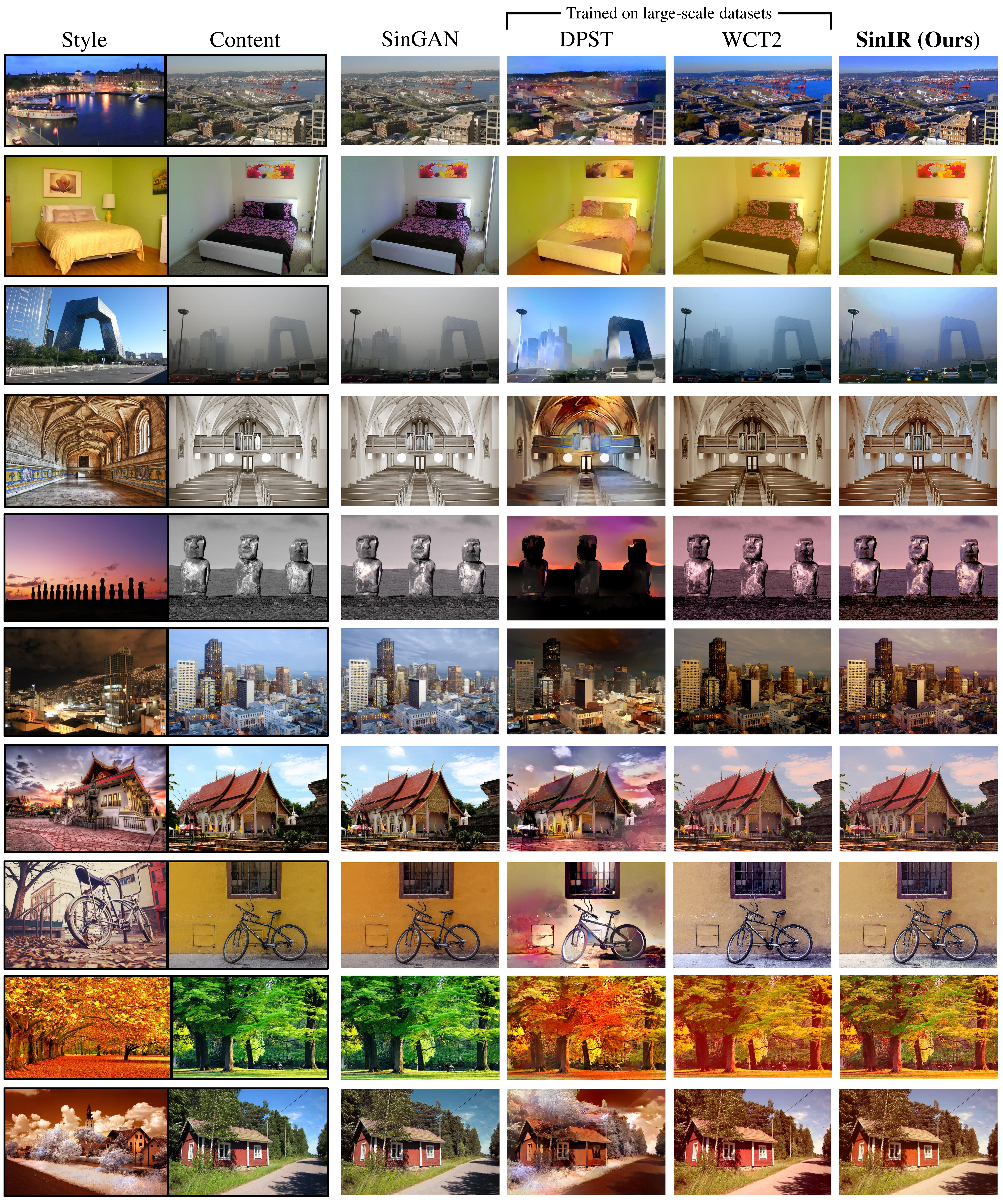}\
    \caption{\textbf{Photo-realistic Style Transfer.} DPST \cite{Luan17} and WCT2 \cite{Yoo19} are dedicated methods.}
\end{figure*}

\begin{figure*}
    \centering
    \includegraphics[width=\textwidth]{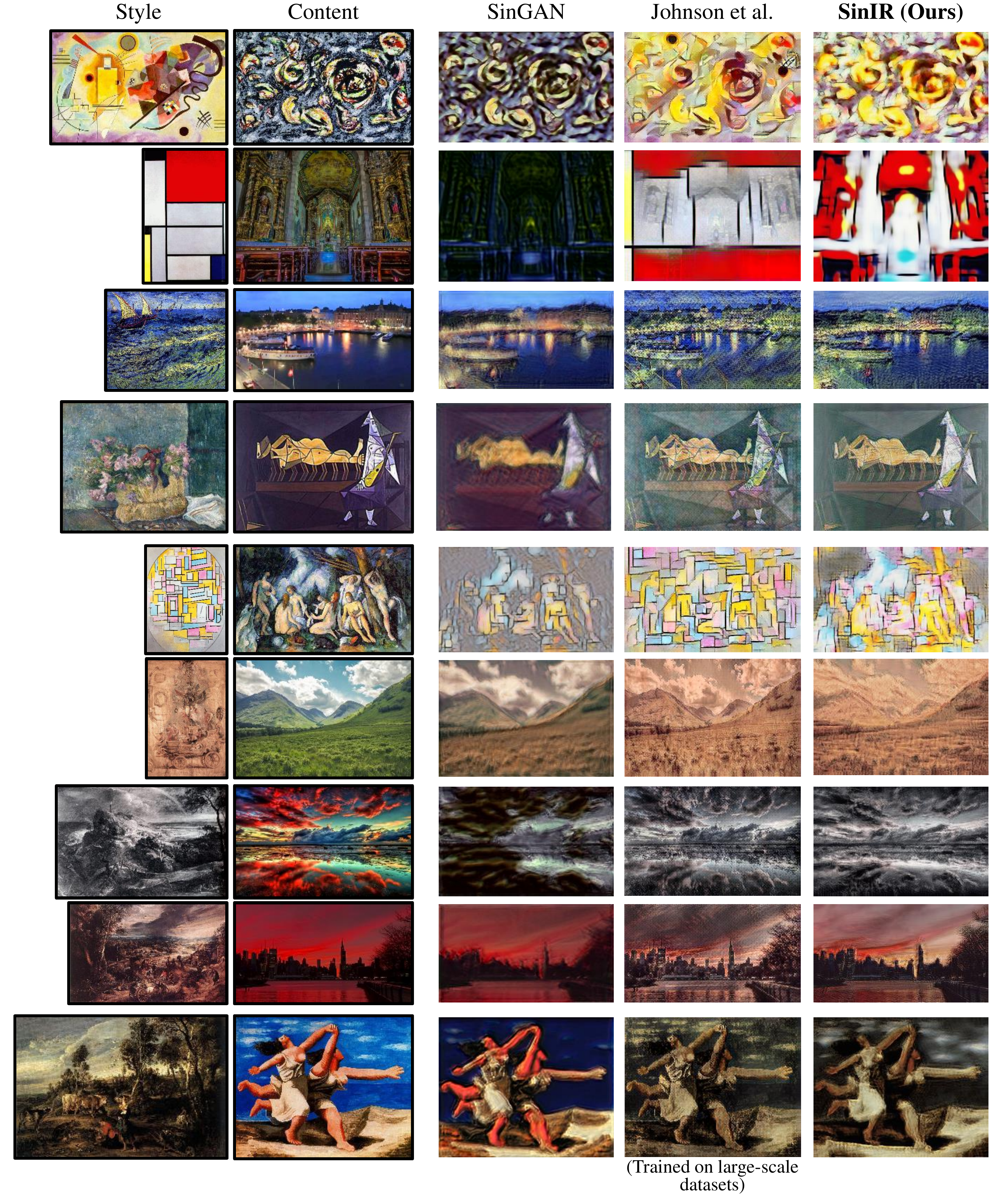}
    \caption{\textbf{Artistic Style Transfer.} \cite{Johnson16} is a dedicated method.}
\end{figure*}

\begin{figure*}
    \centering
    \includegraphics[width=\textwidth]{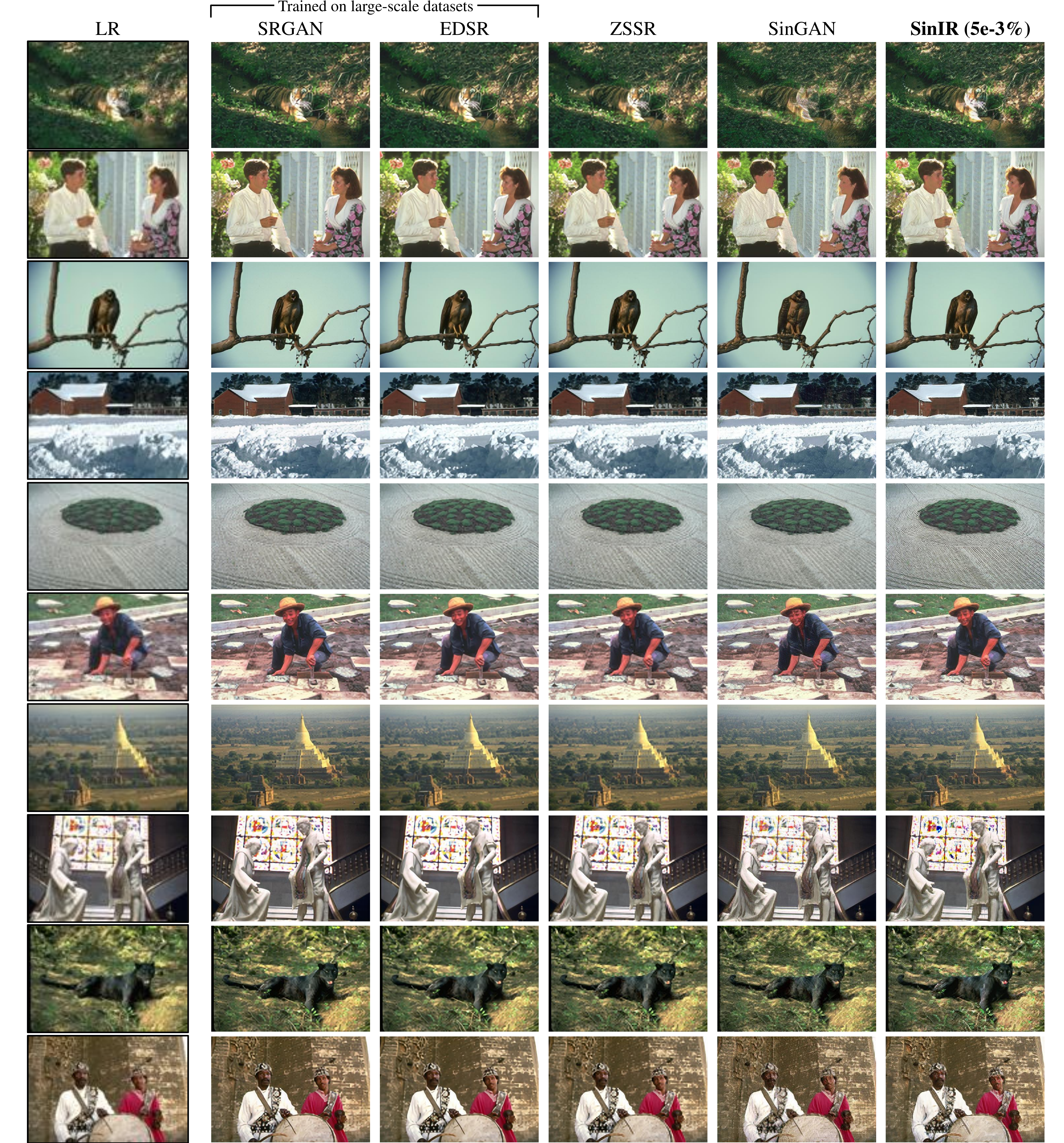}
    \caption{\textbf{Super-resolution.} SRGAN \cite{Ledig17}, EDSR \cite{Lim17}, ZSSR \cite{Shocher18zeroshot} are dedicated methods. For simplicity, only results from SinIR with 5e-3\% of random pixel shuffling are shown. (Please see super-resolution section in the original paper for details.)}
\end{figure*}

\begin{figure*}
    \centering
    \includegraphics[width=\textwidth]{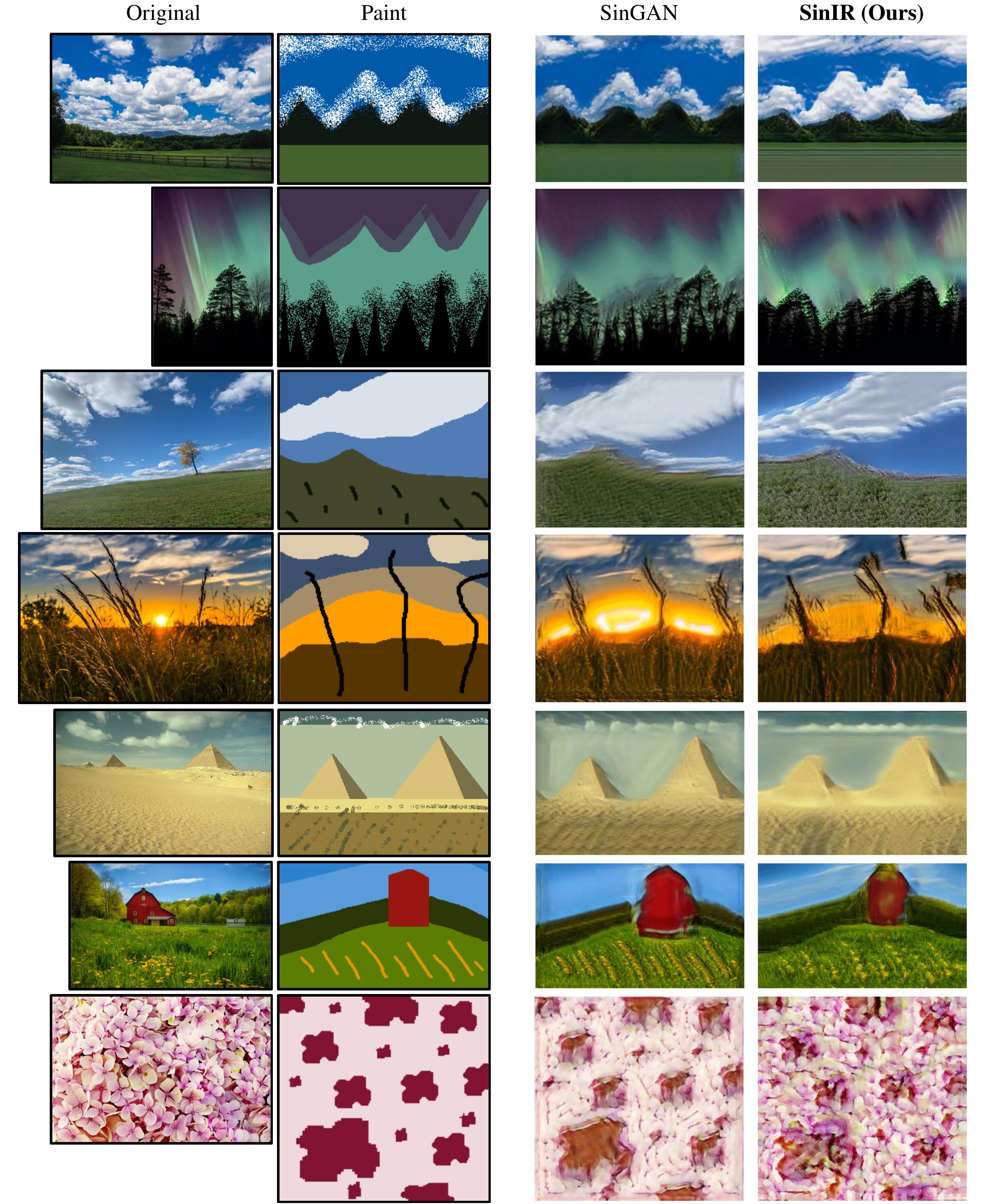}
    \caption{\textbf{Paint-to-Image.}}
\end{figure*}

\begin{figure*}
    \centering
    \includegraphics[width=\textwidth]{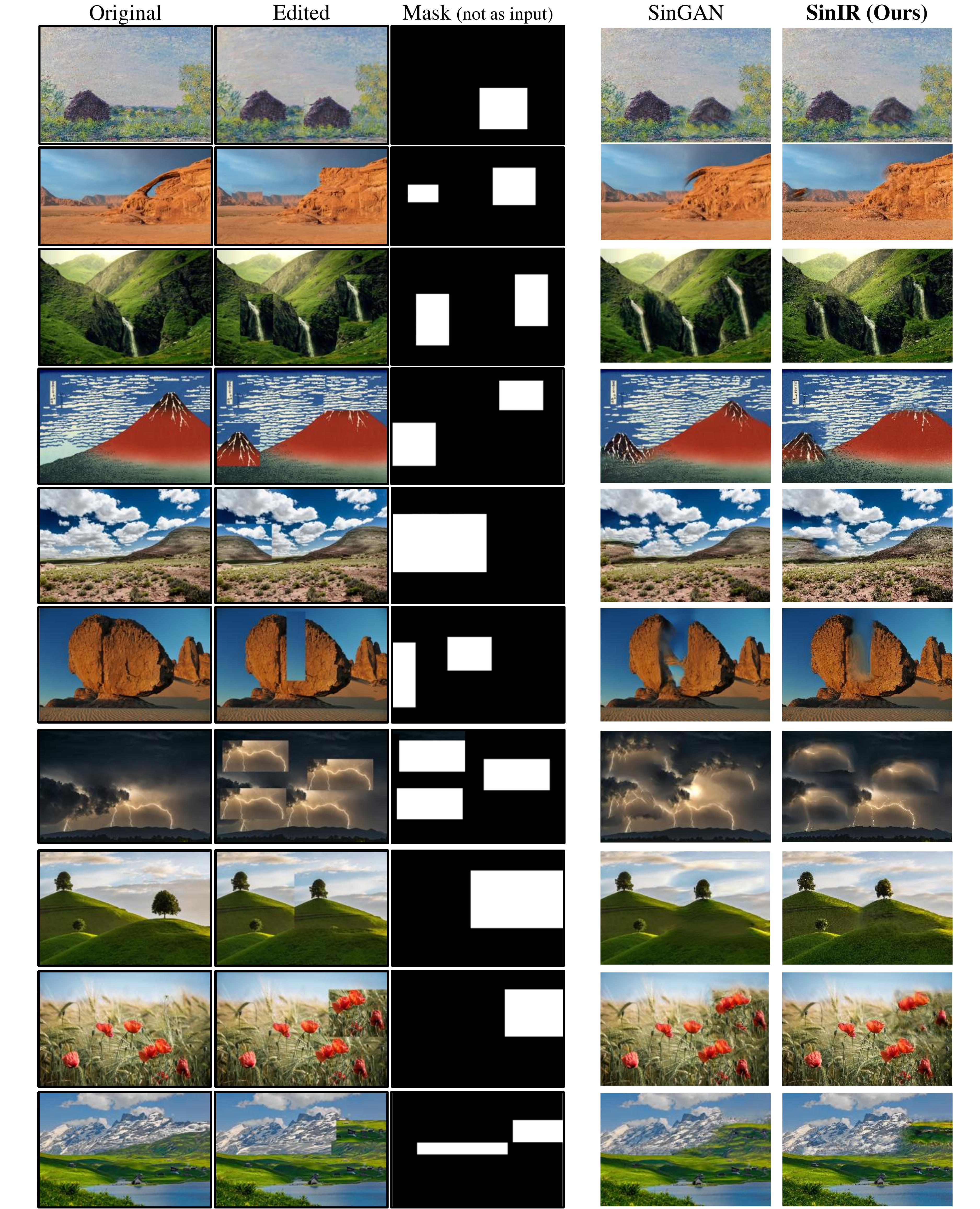}
    \caption{\textbf{Editing.}}
\end{figure*}

\begin{figure*}
    \centering
    \includegraphics[width=\textwidth]{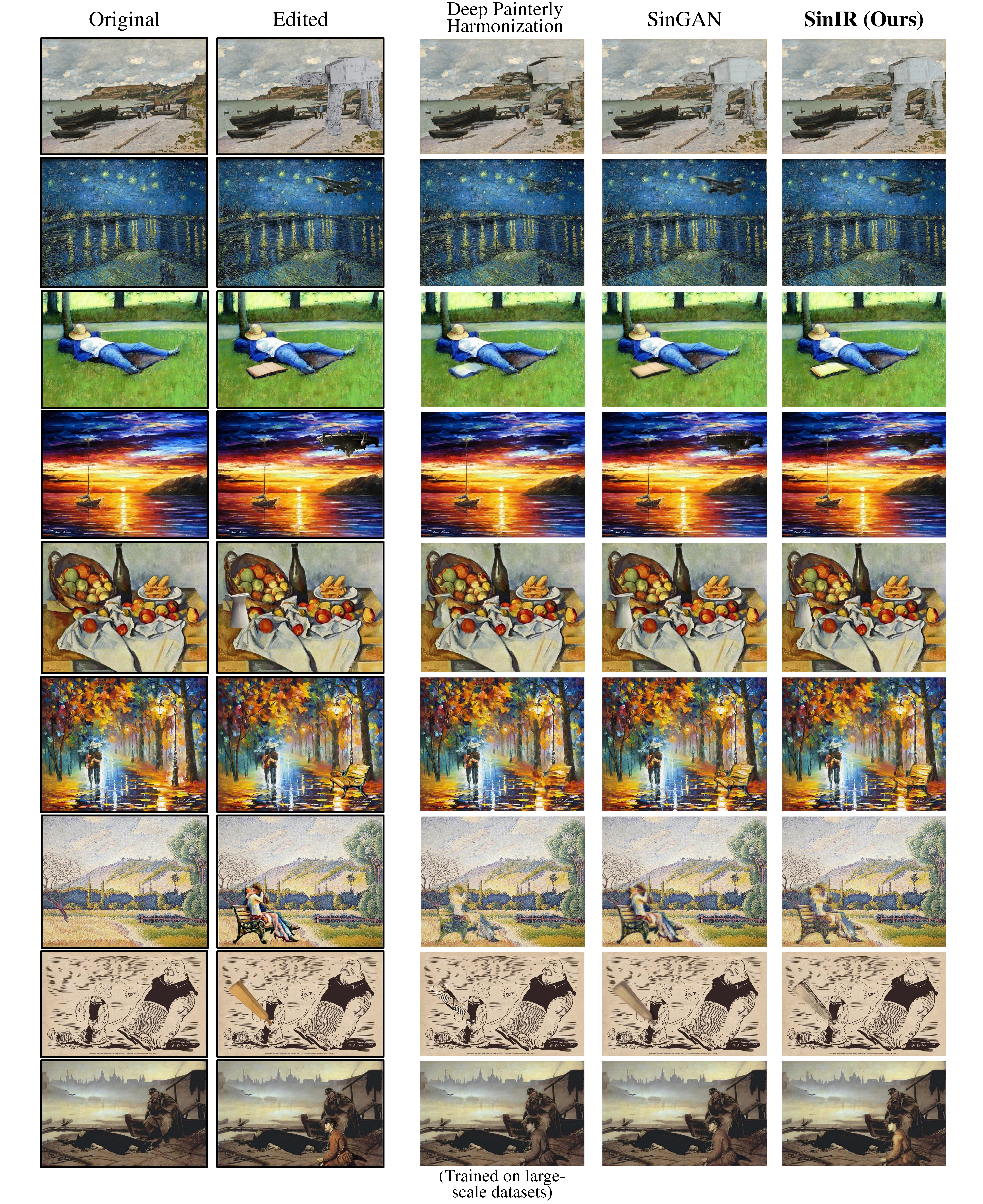}
    \caption{\textbf{Harmonization.} Deep Painterly Harmonization \cite{Luan18} is a dedicated method.}
\end{figure*}

\begin{figure*}[t]
    \centering
    \includegraphics[width=\textwidth]{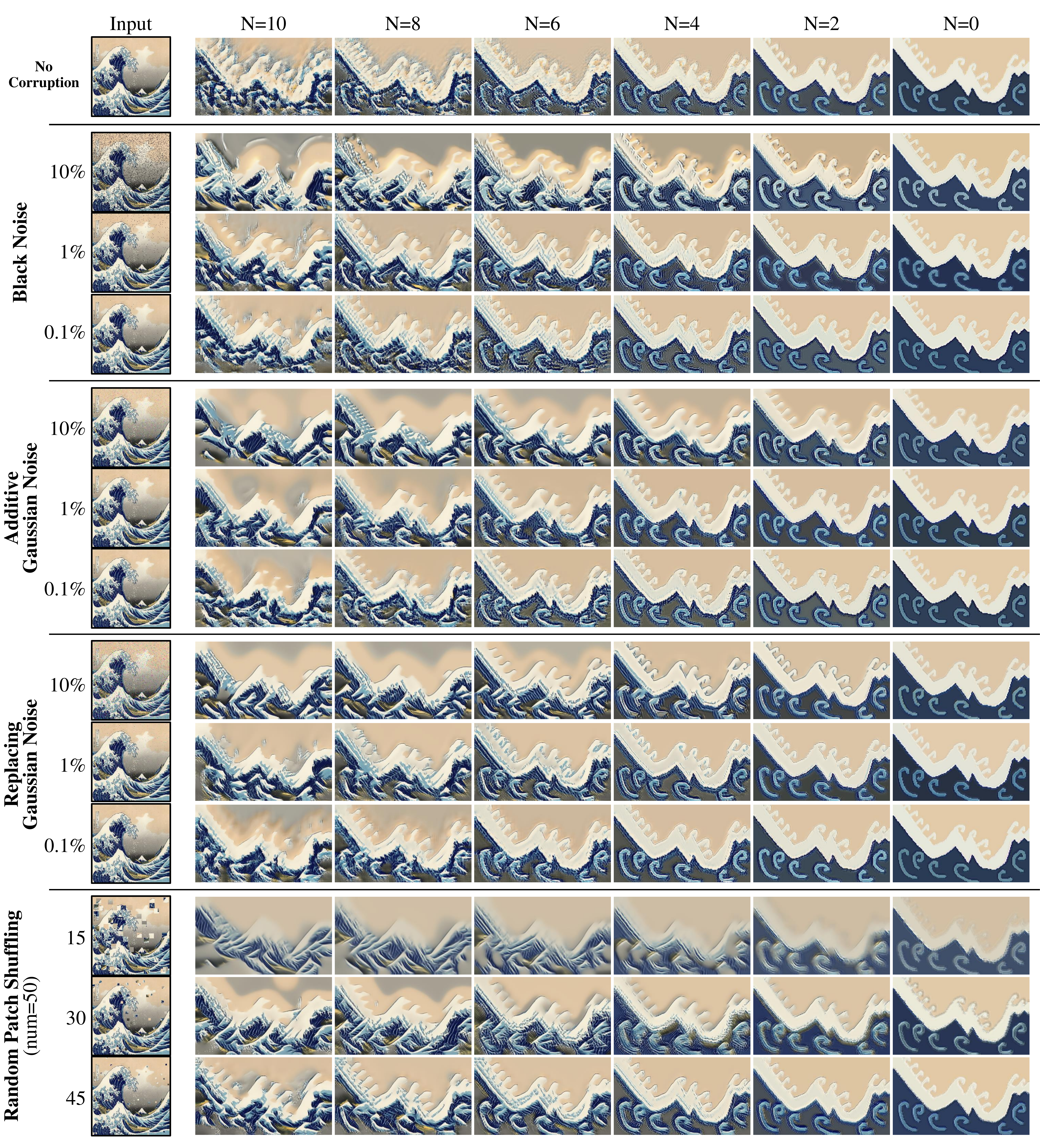}
    \caption{\textbf{Effect of different corrupting noise.} We explore 4 types of corruption. The numbers at the top indicate inference starting scales. Please see Section 2.1 of the original paper. The numbers on the left-hand side of inputs indicate the intensity of corruption. For random patch shuffling, they are the ratio of (a longer side of the original image) to (a side of a patch). For example, in case of the 30, (a longer side of the original image) / 30 = (a side of a patch). For other corruption schemes, the numbers are the percentage of randomly sampled pixels to be corrupted. Please see Section \ref{corrupt_sec} for details.}
    \label{corruption}
\end{figure*}

\clearpage

\section{\label{corrupt_sec}Effect of Different Corrupting Scheme}
Although we use only \emph{random pixel shuffling} in the original paper, alternative corrupting schemes can be considered. Figure \ref{corruption} illustrates the effect of different corruption. Here we explore 4 types of corruption. From Figure \ref{corruption}, we can see that applying mild corruption to the input of SinIR generally gives better results regardless of corruption schemes, compared to those of no corruption (the first row in Figure \ref{corruption}). Also, consistent with the findings discussed in Section 2.1.2 of the original paper, regardless of the corrupting schemes, when the intensity of the corruption becomes high, the results become smoothed and vice versa.

\paragraph{Black Noise}
\emph{Force randomly sampled pixels to be completely black.} This corrupting scheme is originally used in denoising autoencoder \cite{Vincent08} with MNIST dataset \cite{Bengio06, Lecun98}. Compared to other corrupting schemes, this corruption sometimes produces unnatural textures (\emph{e.g.}, floating objects with black edges in Figure \ref{corruption}). This is probably because we are using natural images, not the MNIST dataset. Considering that natural images often include more complex structures, simply \emph{turning off} pixels ignores such visual properties and may prevent learning better relationship between adjacent pixels.

\paragraph{Additive Gaussian Noise}
\emph{Add gaussian noise to randomly sampled pixels.} For gaussian noise, we set mean and variance to 0 and 0.5 with the pixel value of [-1, 1]. The corrupted pixel values are clipped at -1 and 1. Compared to \emph{Black Noise}, this corrupting scheme generally produces better results that are close to random pixel shuffling that is used in the original paper. A possible reason is that now we are corrupting the input based on its original pixel values.

\paragraph{Replacing Gaussian Noise}
\emph{Replace randomly sampled pixels with gaussian noise.} For gaussian noise, we set mean and variance to 0 and 0.5 with the pixel value of [-1, 1]. The corrupted pixel values are clipped at -1 and 1. The results are similar to \emph{Additive Gaussian}, but it sometimes produces unnatural objects as \emph{Black Noise} does. It is probably because both of them corrupt pixels not based on its original values.

\paragraph{Random Patch Shuffling}
\emph{Shuffle randomly sampled patches.} We shuffle 50 patches for Figure \ref{corruption}. As we directly use internal patches, this scheme produces well-textured results (\emph{e.g.} wave bubbles). These results are closest to those of \emph{random pixel shuffling} used in the original paper.
\vspace{1cm}
\end{appendices}
\end{document}